\documentclass[10pt,twocolumn,letterpaper]{article}

\usepackage[pagenumbers]{cvpr} %

\usepackage{graphicx}
\usepackage{amsmath}
\usepackage{amssymb}
\usepackage{booktabs}
\usepackage[table]{xcolor}
\usepackage{tabulary,multirow,overpic,xcolor}

\newcommand{\pool}{\mathcal{P}}

\newcommand{\figref}[1]{Fig.~\ref{fig:#1}}
\newcommand{\tabref}[1]{Table~\ref{tab:#1}}
\newcommand{\secref}[1]{\S\ref{sec:#1}}
\newcommand{\algref}[1]{Algorithm~\ref{alg:#1}}

\usepackage{tabulary}

\usepackage{soul}
\definecolor{defaultcolor}{gray}{.92}
\definecolor{defaultcolor2}{gray}{.90}
\sethlcolor{defaultcolor2}

\newcommand{\rbr}[1]{\left(#1\right)}
\newcommand{\sbr}[1]{\left[#1\right]}

\newcommand{\app}{\raise.17ex\hbox{$\scriptstyle\sim$}}
\def\x{$\times$}

\definecolor{carmine}{rgb}{0.59, 0.0, 0.09}

\newcommand{\dummy}{ {\color{white}(+0.0)}}
\newcommand{\gain}[1]{{\color{chameleon3}(+{#1})}}

\definecolor{demphcolor}{RGB}{144,144,144}
\newcommand{\demph}[1]{\textcolor{demphcolor}{#1}}

\definecolor{xycolor}{RGB}{60, 120, 216}
\definecolor{xycolor}{HTML}{0071bc}
\newcommand{\xycolor}[1]{\textcolor{xycolor}{#1}}
\definecolor{wcolor}{RGB}{103, 78, 167}
\newcommand{\wcolor}[1]{\textcolor{wcolor}{#1}}
\definecolor{dcolor}{RGB}{166, 77,21}

\definecolor{gcolor}{RGB}{204, 102, 153}

\definecolor{tcolor}{RGB}{80, 200, 180}
\newcommand{\tcolor}[1]{\textcolor{citecolor}{#1}}
\definecolor{eicolor}{RGB}{153, 51, 102}
\newcommand{\eicolor}[1]{\textcolor{eicolor}{#1}}
\newcommand{\outsizesRaw}[4]{\multirow{#4}{*}{\(\begin{array}{c}  \text{#1$\times$#2\x #3}\\[-.1em]  \end{array}\)}}
\newcommand{\outsizesRawD}[5]{\multirow{#5}{*}{\(\begin{array}{c}  \text{#1\x#2\x#3\x#4}\\[-.1em]  \end{array}\)}}

\newcommand{\blockatta}[3]{\multirow{2}{*}{\(\left[\begin{array}{c}\text{\eicolor{MHPA}(\wcolor{#1})}\\[-.1em] \text{MLP(\wcolor{#2})}\end{array}\right]\)$\times$#3}
}

\newcolumntype{x}[1]{>{\centering\arraybackslash}p{#1pt}}
\newcolumntype{y}[1]{>{\raggedright\arraybackslash}p{#1pt}}
\newcolumntype{z}[1]{>{\raggedleft\arraybackslash}p{#1pt}}
\newlength\savewidth\newcommand\shline{\noalign{\global\savewidth\arrayrulewidth
		\global\arrayrulewidth 1pt}\hline\noalign{\global\arrayrulewidth\savewidth}}
\newcommand{\tablestyle}[2]{\setlength{\tabcolsep}{#1}\renewcommand{\arraystretch}{#2}\centering\footnotesize}

\usepackage{algorithm}
\usepackage{listings}

\usepackage{etoolbox}
\makeatletter
\AfterEndEnvironment{algorithm}{\let\@algcomment\relax}
\AtEndEnvironment{algorithm}{\kern2pt\hrule\relax\vskip3pt\@algcomment}
\let\@algcomment\relax
\newcommand\algcomment[1]{\def\@algcomment{\footnotesize#1}}
\renewcommand\fs@ruled{\def\@fs@cfont{\bfseries}\let\@fs@capt\floatc@ruled
  \def\@fs@pre{\hrule height.8pt depth0pt \kern2pt}%
  \def\@fs@post{}%
  \def\@fs@mid{\kern2pt\hrule\kern2pt}%
  \let\@fs@iftopcapt\iftrue}
\makeatother
\makeatletter
\@namedef{ver@everyshi.sty}{}
\makeatother

\usepackage{tikz,pgfplots}
\usepackage{tango}

\pgfplotsset{compat = 1.3,
	legend style={font=\scriptsize},
	legend cell align={left},
	legend style={cells={align=left}, draw=black!20},
	grid=both,
	grid style={dotted},
	tick style={draw=none},
	enlarge x limits=false,
	enlarge y limits=false,
	axis line style={draw=black!100},
	axis lines=left,
}

\pgfplotscreateplotcyclelist{mycolormarklist}{%
	chameleon3,mark=diamond*,mark options={solid, fill=chameleon3}\\
	skyblue2,mark=pentagon*,mark options={solid}\\
	scarletred3,mark=triangle*,mark options={solid}\\
	plum1,mark=square*,mark options={solid}\\
	aluminium5,mark=otimes*,mark options={solid}\\
	orange1,mark=square*,mark options={solid}\\
	skyblue3\\
	chocolate2\\
	maroon\\
	butter3\\
}
\pgfplotscreateplotcyclelist{mystylelist}{%
	densely dashdotted\\
	solid\\
	dotted\\
	densely dashed\\
	dashed\\
	dashdotted\\
	densely dotted\\
	densely dash dot dot\\
	dotted\\
	dash dot dot\\
}

\definecolor{citecolor}{RGB}{34,139,34}
\definecolor{citecolor2}{HTML}{0071bc}
\definecolor{lightred}{RGB}{241,140,142}
\usepackage[pagebackref=true,breaklinks=true,letterpaper=true,colorlinks,
citecolor=citecolor2,bookmarks=false]{hyperref}

\usepackage[capitalize]{cleveref}
\crefname{section}{Sec.}{Secs.}
\Crefname{section}{Section}{Sections}
\Crefname{table}{Table}{Tables}
\crefname{table}{Tab.}{Tabs.}

\def\cvprPaperID{***} %
\def\confName{CVPR}
\def\confYear{2022}

\begin{document}

	\title{MeMViT: Memory-Augmented Multiscale Vision Transformer\\for Efficient Long-Term Video Recognition}
\author{
	Chao-Yuan Wu\textsuperscript{ *, 1} \qquad
	Yanghao Li\textsuperscript{ *, 1} \qquad
	Karttikeya Mangalam\textsuperscript{ 1, 2} \\
	Haoqi Fan\textsuperscript{ 1} \qquad 
	Bo Xiong\textsuperscript{ 1}\qquad
	Jitendra Malik\textsuperscript{ 1, 2} \qquad
	Christoph Feichtenhofer\textsuperscript{ *, 1}\\
	\small $^{*}$equal technical contribution   \vspace{.5em} \\
	\textsuperscript{1}Facebook AI Research \qquad \qquad \textsuperscript{2}UC Berkeley 
	 \vspace{-.2em}
}

	\maketitle

\begin{abstract}
While today's video recognition systems parse snapshots or short clips accurately,
they cannot connect the dots and reason across a longer range of time yet.
Most existing video architectures can only process $<$5 seconds of a video without hitting the computation or memory bottlenecks.

In this paper, we propose a new strategy to overcome this challenge.
Instead of trying to process more frames at once like most existing methods,
we propose to process videos in an online fashion and cache ``\emph{memory}" at each iteration.
Through the memory, the model can reference prior context for long-term modeling, with only a marginal cost.
Based on this idea, we build MeMViT, a Memory-augmented Multiscale Vision Transformer,
that has a temporal support 30\x longer than existing models with only 4.5\% more compute; traditional methods need $>$3,000\% more compute to do the same.
On a wide range of settings, the increased temporal support enabled by MeMViT brings large gains in recognition accuracy consistently.
MeMViT obtains state-of-the-art results on the AVA, EPIC-Kitchens-100 action classification, and action anticipation datasets. 
Code and models are available at \url{https://github.com/facebookresearch/memvit}.
\vspace{-4mm}
\end{abstract}
\section{Introduction}
\label{sec:intro}

Our world evolves endlessly over time.
The events at different points in time influence each other and all together, they tell the story of our visual world.
Computer vision promises to understand this story, but today's systems are still quite limited.
They accurately parse visual content in independent snapshots or short time periods (\eg, 5 seconds), but not beyond that.
So, how can we enable accurate long-term visual understanding?
There are certainly many challenges ahead,
but having a model that \emph{practically} runs on \emph{long} videos is arguably an important first step.

\begin{figure}[t]
\centering
\subfloat[\textbf{Traditional long-term models \vs our method, \textbf{MeMViT}.}\label{fig:teaser:a}]{%
\includegraphics[width=\linewidth]{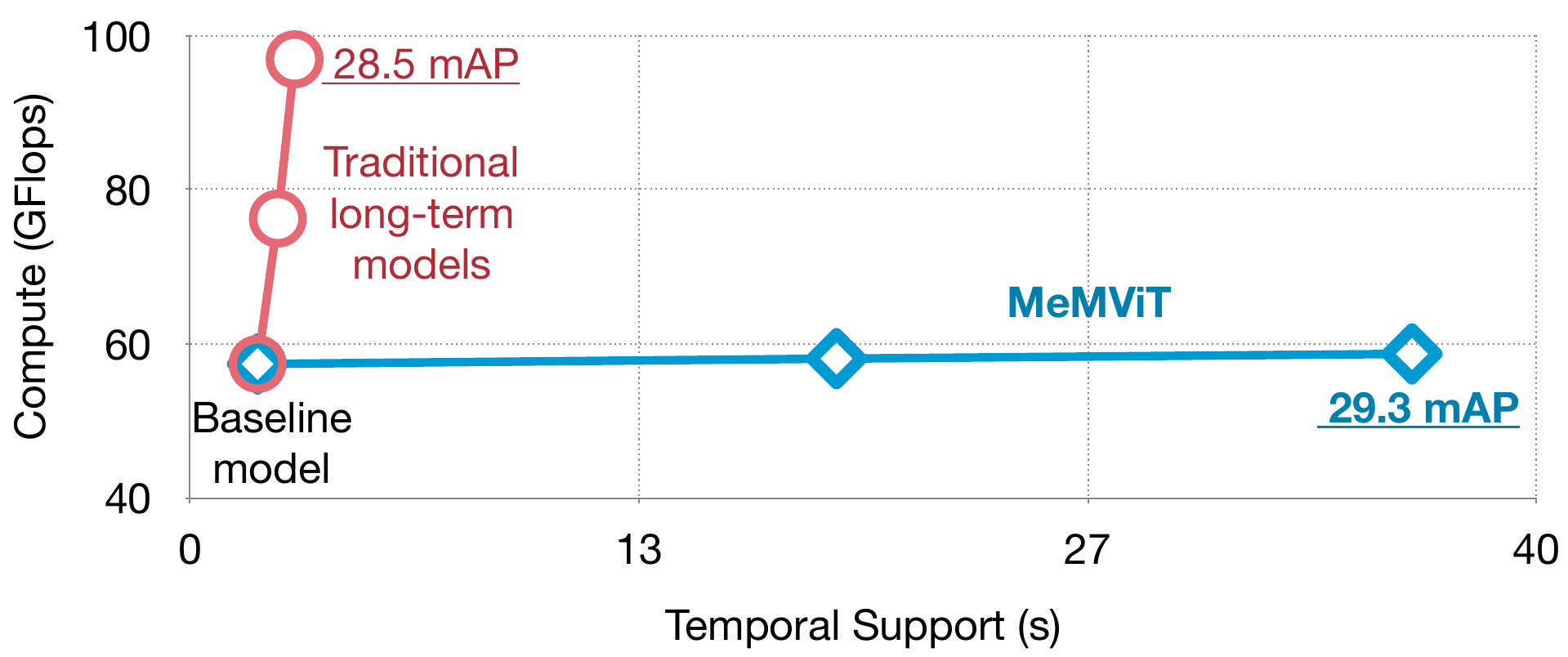}
}
\vspace{3.5mm}

\subfloat[\textbf{MeMViT}\label{fig:teaser:b}]{%
\includegraphics[width=1.0\linewidth]{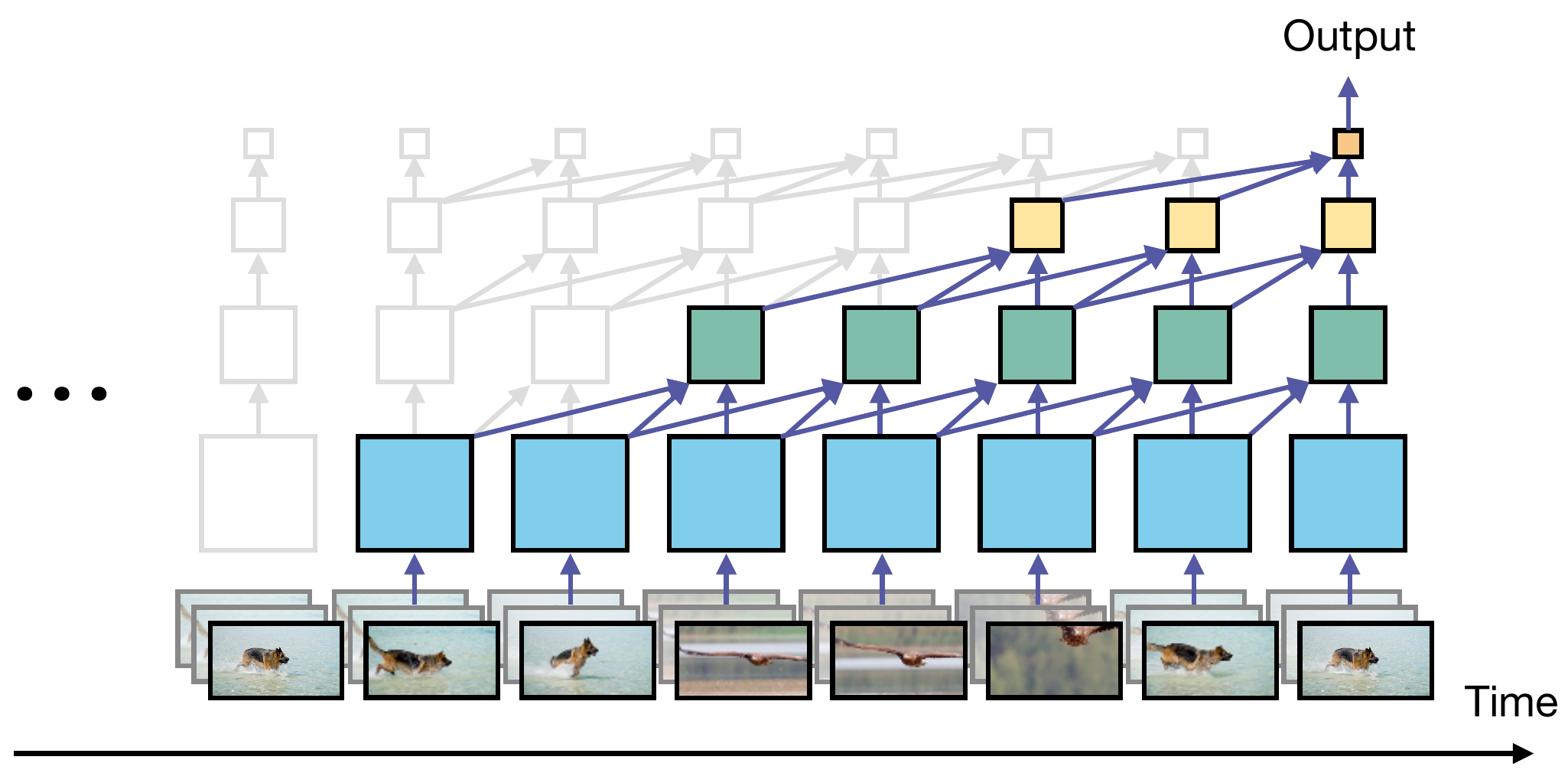}
}
\caption{\textbf{MeMViT} is a class of video models that models long videos efficiently.
It has a significantly better trade-off than traditional methods, which increase the temporal support of a video model by increasing the number of frames in model input (\figref{teaser:a}).
MeMViT achieves efficient long-term modeling by \textit{hierarchically attending} the previously cached ``memory" of the past (\figref{teaser:b}).}
\label{fig:teaser}
\vspace{-3mm}
\end{figure}

In this paper, 
we propose a memory-based approach for building efficient long-term models.
The central idea is that instead of aiming to jointly process or train on the whole long video,
we simply maintain ``memory" as we process a video \emph{in an online fashion}.
At any point of time, the model has access to prior memory for long-term context.
Since the memory is `reused' from the past, the model is highly efficient.
To implement this idea, we build a concrete model called \emph{MeMViT}, a Memory-augmented Multiscale Vision Transformer.
MeMViT processes 30\x~longer input duration than existing models, with only 4.5\% more compute.
In comparison, a long-term model built by increasing the number of frames will require $>$3,000\% more compute.
\figref{teaser:a} presents the trade-off comparison in compute/duration.

More concretely, MeMViT uses the ``keys" and ``values" of a transformer~\cite{vaswani2017attention} as memory.
When the model runs on one clip, the ``queries" attend to an extended set of ``keys" and ``values", which come from both the current time and the past.
When performing this at multiple layers, each layer attends further down into the past,
resulting in a significantly longer receptive field, as illustrated in \figref{teaser:b}.

To further improve the efficiency, 
we jointly train a \emph{memory compression module} for reducing the memory footprint. 
Intuitively, this allows the model to learn which cues are important for {future} recognition and keeps only those.

Our design is loosely inspired by how humans parse long-term visual signals.
Humans do not process all signals over a long period of time at once.
Instead, humans process signals in an \emph{online} fashion, associate what we see to past memory to make sense of it, and also memorize important information for future use. %

Our results demonstrate that 
augmenting video models with memory and enabling long range attention is simple and very beneficial.
On the AVA spatiotemporal action localization~\cite{Gu2018}, the EPIC-Kitchens-100\footnote{The EPIC-Kitchens-100 dataset is licensed under the Creative Commons Attribution-NonCommercial 4.0 International License.} action classification~\cite{Damen2018EPICKITCHENS,Damen2021PAMI}, and the EPIC-Kitchens-100 action anticipation datasets~\cite{Damen2018EPICKITCHENS,Damen2021PAMI},
MeMViT obtains large performance gains over its short-term counterpart and achieves state-of-the-art results.
We hope these results are helpful for the community and take us one step closer to understanding the interesting long story told by our visual world.

\section{Related Work}
\label{sec:related}

\paragraph{Video understanding models} 
aim to parse spatiotemporal information in videos. 
Popular approaches in the past decade include the classic works that use handcrafted features \cite{efros,dollar2005behavior,klaser2008spatio,Dalal2005,Laptev2008,wang2009evaluation,Wang2013,Wang2013a,Peng2014},
recurrent networks~\cite{jiang2019stm, Donahue2015, Ng2015, Li2018a,Li2018,Sun2017}, and 2D-~\cite{Wang2016a,Wu2018,Wang2015} or 3D-CNNs~\cite{Taylor2010, Tran2015, Carreira2018, Feichtenhofer2019, feichtenhofer2020x3d, Zhou2017, Li2018, Tran2019, Qiu2017, Li2018, Xie2018, Wang2018,girdhar2019video}.
More recently, methods built upon the Transformer~\cite{vaswani2017attention} architecture (the vision transformers) have been shown promising results~\cite{MViT,arnab2021vivit,MotionFormer, neimark2021video,bertasius2021space}.

\vspace{-1mm}
\paragraph{Vision transformers}~\cite{ViT, MViT, Swin, arnab2021vivit,dong2021cswin,deit,yuan2021tokens,touvron2021going,Graham_2021_ICCV}
treat an image as a set of patches and model their interactions with transformer-based architectures~\cite{vaswani2017attention}. Recent works adding vision priors such as multi-scale feature hierarchies~\cite{MViT,Graham_2021_ICCV,Swin,wang2021pyramid,yuan2021tokens} or local structure modeling~\cite{Swin,dong2021cswin,chen2021visformer} have shown to be effective. 
They have also been generalized from the image to video domain~\cite{MViT,bertasius2021space,neimark2021video, MotionFormer}.
In this work, we build our architecture based on the Multiscale Vision Transformer (MViT) architecture~\cite{MViT,li2021improved} as a concrete instance, 
but the general idea can be applied to other ViT-based video models.

\vspace{-1mm}
\paragraph{Long-term video models}
aim to capture longer-term patterns in long videos (\eg, $>$30 seconds).
To reduce the high computational cost,
one widely studied line of work directly models pre-computed features without jointly training backbones~\cite{Abu-El-Haija2016,yue2015beyond,wu2021towards,Donahue2015,girdhar2017actionvlad}.
Another potential direction designs efficient models~\cite{Zolfaghari2018,korbar2019scsampler,Wu2018,lin2019tsm,hussein2019timeception,Zhou2017} to make covering more frames feasible.
More related to our work is the less-studied middle ground that builds a memory-like design that still allows for end-to-end training but has greatly reduced overhead~\cite{Wu2019,lee2018memory,lee2021video,chen2020memory}. 
For example, `long-term feature bank'-based methods extend standard video backbones to reference long-term supportive context features~\cite{Wu2019,pan2021actor}.
However, these methods capture only final-layer features and require two backbones, two rounds of training and inference computation.
MeMViT flexibly models features at arbitrary layers with minimal changes to standard training methods and only requires one standalone backbone.

\vspace{-1mm}
\paragraph{Online video modeling} arises naturally in applications such as robotics, AR/VR, or video streaming.
While one may use an image-based method (\eg, \cite{Ren2015}) to parse a video frame-by-frame, to consider longer-term context, most existing works use causal convolutions~\cite{Carreira_2018_ECCV,kondratyuk2021movinets,cheng2019sparse}, RNNs~\cite{Donahue2015,liu2018mobile}, or feature fusion~\cite{zhu17fgfa,chen2020memory}. 
In this work, we explore attention-based designs,
which directly reference arbitrary points of time in the past, without the need to fight forgetfulness as in RNNs or being constrained by kernel size as in CNNs.

\vspace{-1mm}
\paragraph{Transformer designs in NLP}
are also related to our method.
MeMViT takes inspiration from long-range language models~\cite{dai2019transformer,rae2019compressive,sukhbaatar2019adaptive,sukhbaatar2021not,rae2020transformers}, which also cache long-range ``memory".
Different from these works, video models process significantly larger tensors ($T{\times}W{\times}H$), making caching and attending memory expensive if not prohibitive.
Prior work in NLP attempts to learn a module to compress memory, but the requirement of backpropagation through time (BPTT) makes it challenging~\cite{rae2019compressive}.
Rae~\etal~\cite{rae2019compressive} thus uses autoencoder for memory compression, but that cannot be optimized for the end task.
In this paper, we present a ``pipelined" memory compression method that is efficient and end-to-end optimizable for the end task, without BPTT.

\begin{figure*}[t]
	\centering
	\includegraphics[width=\textwidth]{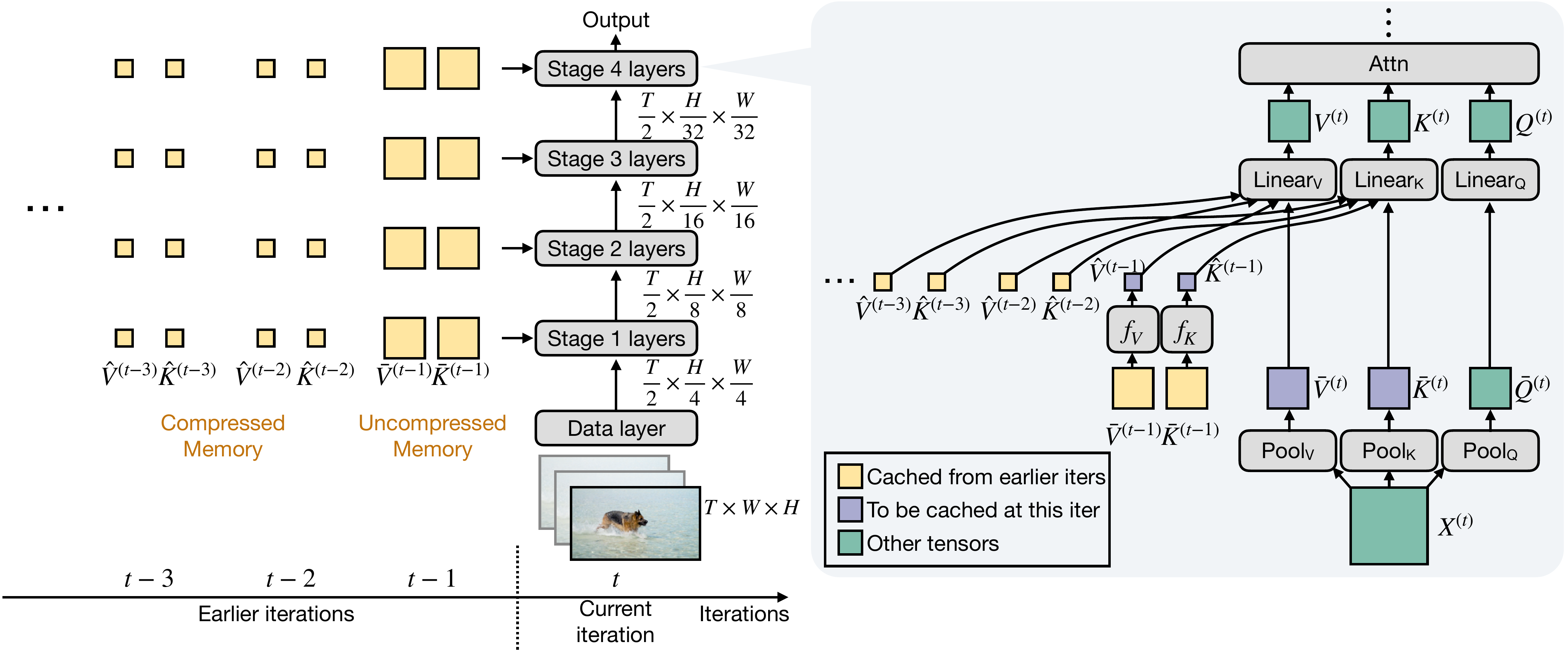}
	\caption{\textbf{MeMViT}
	is a memory-augmented mulstiscale vision transformer network for long-term video recognition. 
	MeMViT treats a long video as a sequence of short clips and process them \emph{sequentially}. (Consecutive iterations see consecutive clips.)
	``Memory" obtained from earlier iterations were cached, so that MeMViT processing the current clip can reference the memory.
	Note that at the current iteration we cache the \emph{uncompressed} memory, which will only be compressed at the next iteration. See text for details.
	Left: Model overview. Right: Detailed MeMViT attention design.%
	}
	\label{fig:model}
\end{figure*}

\section{Preliminaries}
\label{sec:bg}
In this paper, we build MeMViT based on the MViT~\cite{MViT,li2021improved} architecture due to its strong performance, 
but the techniques presented in this paper can be applied to other ViT-based architectures.
For completeness, we review ViT and MViT and introduce notations used in this paper next.

\paragraph{Vision Transformers (ViT)} first embeds an image into $N$ non-overlapping patches (using a strided convolution) and packs them into a tensor $X^0 \in \mathbb{R}^{N \times d}$.
A stack of transformer layers then models the interactions among these patches.
The central component of a transformer layer is the attention operation, which first linearly projects an input tensor $X$ to be queries $Q$, keys $K$, and values $V$:\footnote{Here we omit the layer index for clarity.}
\begin{align}
Q = X W_Q, \quad K = X W_K, \quad V = X W_V,
\end{align}
and performs a self-attention operation
\begin{align}
Z := \mathrm{Attn}(Q, K, V) = \mathrm{Softmax}\rbr{QK^\top / \sqrt{d}}V,
\end{align}
to obtain an output tensor $Z \in \mathbb{R}^{N \times d_{\textrm{out}}}$.

\paragraph{Multiscale Vision Transformers (MViT)}
improves ViT based on two simple ideas.
First, 
instead of having a fixed resolution of $N$ throughout the network, 
MViT learns multi-scale representations through multiple \emph{stages}, starting from fine-grained modeling of smaller patches (with large $N$ and small $d$) to high-level modeling of larger patches in later stages (with small $N$ and large $d$).
The transition between stages is done through strided pooling.
Second, MViT uses \emph{pooling attention} ($\pool$) that pools spatiotemporal dimensions of $Q$, $K$, and $V$ to drastically reduce computational cost of an attention layer, \ie,
\begin{align*}
Q = \pool_Q\rbr{X W_Q}, \  K = \pool_K\rbr{X W_K}, \ V = \pool_V\rbr{X W_V}.
\end{align*}

These two changes significantly improve the model performance and efficiency.
In this paper, we build our method based on a slightly modified MViT,
where we swap the order of linear layer and pooing:
\begin{align}
\bar{Q} = \pool_Q\rbr{X}, \quad  \bar{K} = \pool_K\rbr{X}, \quad \bar{V} = \pool_V\rbr{X}\\
Q = \bar{Q} W_Q, \quad  K = \bar{K} W_K, \quad V = \bar{V} W_V\quad
\end{align}
This allows the linear layer to operate on smaller tensors, reducing the computational cost without affecting accuracy.
See the Appendix for an ablation study on this change.
In the next section, we will see how this change also makes MeMViT more efficient.

To build longer duration video models, 
most state-of-the-art methods simply increase the number of frames in the input clip~\cite{Wang2018,Feichtenhofer2019,MViT}.
This strategy increases the computational cost significantly.
In the next section, we present our method for building more efficient long-term video models.

\section{MeMViT for Efficient Long-Term Modeling}
\label{sec:main}
Our method is simple.
We split a video into a sequence of short $T\times H\times W$ clips and process them \emph{sequentially} (for both training and inference).
Consecutive iterations see consecutive clips.
We cache ``memory", some representations of the processed clip, at each iteration. 
When processing the current clip at time step $t$, the model has access to previously cached `memory' from earlier iterations $t'<t$ for long-term context.
\figref{model} shows an overview.

\subsection{Memory Attention and Caching}

\paragraph{The Basic MeMViT Attention.}
One simple way to implement this idea is to treat the ``keys" $\bar{K}$ and ``values" $\bar{V}$ in the transformer architecture as a form of memory,
and extend $\bar{K}^{(t)}$ and $\bar{V}^{(t)}$ at current iteration $t$ to include $\bar{K}^{(t')}$ and $\bar{V}^{(t')}$ cached from earlier iterations $t'$ from $t-M$ to $t-1$, \ie,
\begin{align}
\bar{K}^{(t)} &:= \sbr{\mathrm{sg}\rbr{\bar{K}^{(t-M)}}, \ldots, \mathrm{sg}\rbr{\bar{K}^{(t-1)}}, \bar{K}^{(t)}},\\ 
\bar{V}^{(t)} &:= \sbr{\mathrm{sg}\rbr{\bar{V}^{(t-M)}}, \ldots, \mathrm{sg}\rbr{\bar{V}^{(t-1)}}, \bar{V}^{(t)}},
\end{align}
where the square brackets denote concatenation along the token dimension. 
With this formulation, the query $Q$ attends not only information about the current time step $t$, but also information from up to $M$ steps before.\footnote{Note that we operate on $\bar{K}$ and $\bar{V}$ instead of $K$ and $V$ so that the following linear layer will transform the features before the attention operation. In preliminary experiments we find this to perform better.}
Here, the ``stop gradient" operator ($\mathrm{sg}$) breaks further dependency into the past in backpropagation. 
Note that the memory is built \textit{hierarchically over time} (see \figref{teaser:b}) and our previous key and value memory holds information stored from prior time-steps.

The additional cost for training and inference encompasses only the GPU memory for memory caching
and the extra compute in the extended attention layer.
All other parts of the network (MLPs, \etc) remain unchanged.
The cost grows with temporal support in $\mathcal{O}\rbr{M}$, instead of $\mathcal{O}\rbr{T^2}$ as in traditional scaling methods.

In this basic implementation, we cache the \emph{full} key and value tensors, which may contain redundant information that is not useful for future recognition.
In the next section we will discuss methods to compress memory for keeping only `important' information.

\begin{algorithm}[t]
\caption{Pseudocode of MeMViT attention in a PyTorch-like style.}
\label{alg:code}
\algcomment{
\fontsize{7.2pt}{0em}\selectfont 
\texttt{cat}: concatenation along token dimension.
}
\definecolor{codeblue}{rgb}{0.25,0.5,0.5}
\lstset{
  backgroundcolor=\color{white},
  basicstyle=\fontsize{7.2pt}{7.2pt}\ttfamily\selectfont,
  columns=fullflexible,
  breaklines=true,
  captionpos=b,
  commentstyle=\fontsize{7.2pt}{7.2pt}\color{codeblue},
  keywordstyle=\fontsize{7.2pt}{7.2pt},
}
\begin{lstlisting}[language=python]
class MeMViTAttention():
  # pool_q, pool_k, pool_v: pooling layers
  # lin_q, lin_k, lin_v: linear layers
  # f_k, f_v: compression modules

  self.m_k = [] # cached memory keys
  self.m_v = [] # cached memory values
  self.max_len  # max memory length

  def forward(x):
    # compute the pooled Q, K, and V
    q, k, v = pool_q(x), pool_k(x), pool_v(x)

    # compress memory
    cm_k = f_k(m_k[-1])
    cm_v = f_v(m_v[-1])

    # perform attention on augmented keys and values
    z = attn(
      lin_q(q), 
      lin_k(cat(self.m_k[:-1] + [cm_k, k])),
      lin_v(cat(self.m_v[:-1] + [cm_v, v])),
    )

    # cache newly compressed memory
    self.m_k[-1] = cm_k.detach()
    self.m_v[-1] = cm_v.detach()

    # cache current uncompressed memory
    self.m_k.append(k.detach())
    self.m_v.append(v.detach())

    # maintain max length for memory
    if len(self.m_k) > self.max_mem:
       self.m_k.pop_first()
       self.m_v.pop_first()
    return z

\end{lstlisting}
\vspace{2mm}
\end{algorithm}

\subsection{Memory Compression}
\paragraph{Na\"ive Memory Compression.}
There are many potential ways to compress the memory, but one intuitive design attempts to jointly train compression modules (\eg, learnable pooling operators), $f_K$ and $f_V$, to reduce the spatiotemporal size of $K$ and $V$ tensors, respectively:
\begin{align*}
\bar{K}^{(t)}:=\sbr{f_K\rbr{\mathrm{sg}(\bar{K}^{(t-M)})}, \ldots,  f_K\rbr{\mathrm{sg}(\bar{K}^{(t-1)})}, \bar{K}^{(t)}},
\end{align*}
and similarly for $\bar{V}^{(t)}$.
With this design, we only need to cache and attend the `compressed' memory, $f_K\rbr{\bar{K}^{(t')}}$ and $f_V\rbr{\bar{V}^{(t')}}$, at \emph{inference} time,
thus reducing the memory footprint and the computational cost.
Nonetheless, at \emph{training} time, it needs to jointly train on all the `full' memory tensor, thus which
may actually \emph{increase} the memory consumption and cost, making obtaining such a model expensive.
The cost is even higher for models with a larger $M$ for longer-term modeling.\footnote{
We will present more empirical analysis in \secref{ablation}.}

	\begin{figure*}
		\resizebox{1\textwidth}{!}{
			\subfloat[\textbf{}\label{fig:tradeoff:trmem}]{%
				\begin{tikzpicture}
				\begin{axis}[
				clip=false,
				width=0.24\textwidth,
				height=0.24\textwidth,
				xlabel={Temporal support (s)},
				ylabel={Train GPU mem (GB)},
				xmin=0,
				ymax=12,
				every axis plot/.append style={very thick,mark options={scale=0.5, solid}},
				legend style={at={(1.0,1.0)},anchor=north east, cells={align=left}}
				]
				\addplot+[scarletred3,mark=triangle*,mark options={solid}, dotted]
				table[row sep=crcr] {
					x y\\
					2.0	4.15\\
					2.6	5.40\\
					3.1	6.81\\
					3.6	8.36\\
					4.2	10.08\\
				};

				\addplot+[skyblue2,mark=pentagon*,mark options={solid}, solid]
				table[row sep=crcr] {
					x y\\
					2.0	4.15\\
					19.2	4.82\\
					36.3	5.52\\
					53.3	6.22\\
					70.4	6.91\\
				};

				\addplot+[chameleon3,mark=diamond*,mark options={solid, fill=chameleon3}, densely dashdotted]
				table[row sep=crcr] {
					x y\\
					2.0	4.15\\
					19.2	4.29\\
					36.3	4.34\\
					53.3	4.39\\
					70.4	4.42\\
				};
				
				\legend{Baseline, MeMViT w/o\\compress, \textbf{MeMViT}}
				
				\end{axis}
				\end{tikzpicture}
			}\hfill
			\subfloat[\textbf{}\label{fig:tradeoff:temem}]{%
				\begin{tikzpicture}
				\begin{axis}[
				clip=false,
				width=0.24\textwidth,
				height=0.24\textwidth,
				xlabel={Temporal support (s)},
				ylabel={Test GPU mem (GB)},
				xmin=0,
				every axis plot/.append style={very thick,mark options={scale=0.5, solid}},
				legend style={at={(1.0,1.0)},anchor=north east, cells={align=left}}
				]
				\addplot+[scarletred3,mark=triangle*,mark options={solid}, dotted]
				table[row sep=crcr] {
					x y\\
					2.0	2.08\\
					2.6	2.71\\
					3.1	3.40\\
					3.6	4.18\\
					4.2	5.01\\
				};

				\addplot+[skyblue2,mark=pentagon*,mark options={solid}, solid]
				table[row sep=crcr] {
					x y\\
					2.0	2.08\\
					19.2	2.41\\
					36.3	2.70\\
					53.3	2.99\\
					70.4	3.28\\
				};

				\addplot+[chameleon3,mark=diamond*,mark options={solid, fill=chameleon3}, densely dashdotted]
				table[row sep=crcr] {
					x y\\
					2.0	2.08\\
					19.2	2.18\\
					36.3	2.20\\
					53.3	2.22\\
					70.4	2.25\\
				};
				
				\end{axis}
				\end{tikzpicture}
			}\hfill
			\subfloat[\textbf{}\label{fig:tradeoff:trtime}]{%
				\begin{tikzpicture}
				\begin{axis}[
				clip=false,
				width=0.24\textwidth,
				height=0.24\textwidth,
				xlabel={Temporal support (s)},
				ylabel={Train iter time (s)},
				every axis plot/.append style={very thick,mark options={scale=0.5, solid}},
				xmin=0,
				cycle multiindex* list={%
					mycolormarklist
					\nextlist
					mystylelist
				}
				]
				\addplot+[scarletred3,mark=triangle*,mark options={solid}, dotted]
				table[row sep=crcr] {
					x y\\
					2.0	0.92\\
					2.6	1.22\\
					3.1	1.53\\
					3.6	1.83\\
					4.2	2.18\\
				};

				\addplot+[skyblue2,mark=pentagon*,mark options={solid}, solid]
				table[row sep=crcr] {
					x y\\
					2.0	0.92\\
					19.2	0.93\\
					36.3	1.16\\
					53.3	1.31\\
					70.4	1.31\\
				};

				\addplot+[chameleon3,mark=diamond*,mark options={solid, fill=chameleon3}, densely dashdotted]
				table[row sep=crcr] {
					x y\\
					2.0	0.92\\
					19.2	1.00\\
					36.3	1.10\\
					53.3	1.13\\
					70.4	1.12\\
				};
				
				\end{axis}
				\end{tikzpicture}
			}\hfill
			\subfloat[\textbf{}\label{fig:tradeoff:tetime}]{%
				\begin{tikzpicture}
				\begin{axis}[
				clip=false,
				width=0.24\textwidth,
				height=0.24\textwidth,
				xlabel={Temporal support (s)},
				ylabel={Test iter time (s)},
				every axis plot/.append style={very thick,mark options={scale=0.5, solid}},
				xmin=0,
				cycle multiindex* list={%
					mycolormarklist
					\nextlist
					mystylelist
				}
				]
				\addplot+[scarletred3,mark=triangle*,mark options={solid}, dotted]
				table[row sep=crcr] {
					x y\\
					2.0	0.0939\\
					3.1	0.1425\\
					3.6	0.1713\\
					4.2	0.2018\\
					4.7	0.2346\\
					5.2	0.2759\\
				};

				\addplot+[skyblue2,mark=pentagon*,mark options={solid}, solid]
				table[row sep=crcr] {
					x y\\
					2.0	0.0939\\
					19.2	0.1569\\
					36.3	0.1637\\
					53.3	0.1677\\
					70.4	0.1683\\
				};

				\addplot+[chameleon3,mark=diamond*,mark options={solid, fill=chameleon3}, densely dashdotted]
				table[row sep=crcr] {
					x y\\
					2.0	0.0939\\
					19.2	0.1516\\
					36.3	0.1501\\
					53.3	0.1532\\
					70.4	0.1528\\
				};
				
				\end{axis}
				\end{tikzpicture}
			}\hfill
			\subfloat[\textbf{}\label{fig:tradeoff:flops}]{%
				\begin{tikzpicture}
				\begin{axis}[
				clip=false,
				width=0.24\textwidth,
				height=0.24\textwidth,
				xlabel={Temporal support (s)},
				ylabel={GFLOPs},
				xmin=0,
				every axis plot/.append style={very thick,mark options={scale=0.5, solid}},
				cycle multiindex* list={%
					mycolormarklist
					\nextlist
					mystylelist
				}
				]
				\addplot+[scarletred3,mark=triangle*,mark options={solid}, dotted]
				table[row sep=crcr] {
					x y\\
					2.0	57.40\\
					2.6	76.27\\
					3.1	96.95\\
					3.6	119.44\\
					4.2	143.75\\
				};

				\addplot+[skyblue2,mark=pentagon*,mark options={solid}, solid]
				table[row sep=crcr] {
					x y\\
					2.0	57.40\\
					19.2	65.25\\
					36.3	73.00\\
					53.3	80.75\\
					70.4	88.50\\
				};

				\addplot+[chameleon3,mark=diamond*,mark options={solid, fill=chameleon3}, densely dashdotted]
				table[row sep=crcr] {
					x y\\
					2.0	57.40\\
					19.2	58.09\\
					36.3	58.71\\
					53.3	59.33\\
					70.4	59.95\\
				};
				
				\end{axis}
				\end{tikzpicture}
			}\hfill
			\subfloat[\textbf{}\label{fig:tradeoff:map}]{%
				\begin{tikzpicture}
				\begin{axis}[
				clip=false,
				width=0.24\textwidth,
				height=0.24\textwidth,
				xlabel={GFLOPs},
				ylabel={mAP (\%)},
				every axis plot/.append style={very thick,mark options={scale=0.5, solid}},
				cycle multiindex* list={%
					mycolormarklist
					\nextlist
					mystylelist
				}
				]
				 \addplot%
				 table[row sep=crcr] {
				 x y\\
				57.4 27.0\\
				58.1 28.7\\
				58.7 29.3\\
				 };

				 \addplot%
				 table[row sep=crcr] {
				 x y\\
						57.40 27.0\\
						65.25 28.6\\
						73.00 28.9\\
				 };

				 \addplot%
				 table[row sep=crcr] {
				 x y\\
					57.40 27.0\\
					76.27 28.0\\
					96.95 28.5\\
				 };
				\end{axis}
				\end{tikzpicture}
			}
		}\vspace{-2mm}
		\caption{\textbf{Comparison of Scaling Strategies.}
		Scaling with MeMViT obtains significantly better trade-off than alternative strategies in terms of training GPU memory (\figref{tradeoff:trmem}), inference GPU memory (\protect\ref{fig:tradeoff:temem}), training runtime (\protect\ref{fig:tradeoff:trtime}), inference runtime (\protect\ref{fig:tradeoff:tetime}) and FLOPs (\protect\ref{fig:tradeoff:flops}), while being \textbf{more accurate} (\protect\ref{fig:tradeoff:map}).
		(The widely used `baseline scaling' strategy increases the temporal support of a video model by increasing the number of frames $T$ in input.)
		All methods use the same hardware and software implementation.\vspace{-2mm}
		}\label{fig:tradeoff}
	\end{figure*}
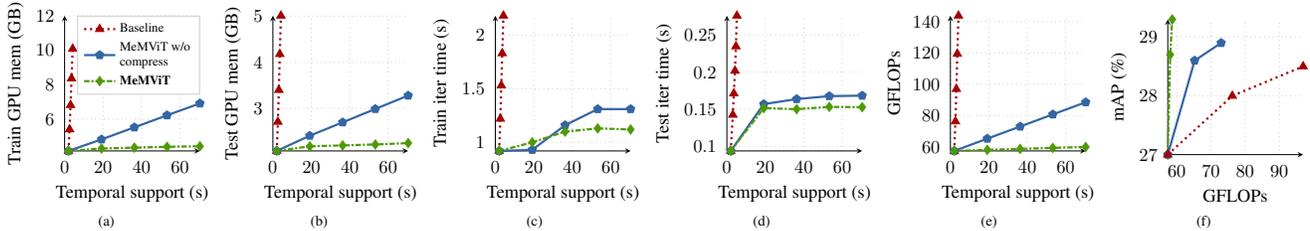

\paragraph{Pipelined Memory Compression.}
To address this issue, we propose a pipelined compression method.
Our insight is that while the compression modules $f_K$ and $f_V$ need to run on uncompressed memory and be jointly optimized,
so that the model learns what is important to keep,
the learned modules can be shared across all the past memory.
Thus, we propose to train to compress memory at only \emph{one step} at a time, \ie,
\begin{align*}
\bar{K}^{(t)} &:= \sbr{\hat{K}^{(t-M)}, \ldots, \hat{K}^{(t-2)}, f_K\rbr{\mathrm{sg}\rbr{\bar{K}^{(t-1)}}}, \bar{K}^{(t)}},
\end{align*}
and similarly for $\bar{V}^{(t)}$.
The right hand side of \figref{model} illustrates this design.
Note that here only the memory $\mathrm{sg}\rbr{\bar{K}^{(t-1)}}$
from the immediate previous step is cached \emph{uncompressed}, and to be used to train $f_K$ in the current iteration.
$\hat{K}^{(t')} = \mathrm{sg}\rbr{f_K(\bar{K}^{(t')})}$ for $t'$ from $t-M$ to $t-2$
are \emph{compressed} memory cached from earlier iterations.
\algref{code} presents the pseudo code for this process.

In this way,
MeMViT adds only a `constant' compression cost over the `basic' MeMViT, since it only runs compression on one single step at a time.
But, it reduces the caching and attention cost for \emph{all other steps} drastically (\eg 16\x~by default).
In \secref{exp}, we will show that overall this leads to significant saving, while maintaining high accuracy.

One appealing property of our design is that the receptive field of our video models grows not only with $M$ but also the number of layers $L$, since each layer attends further down into the past, therefore hierarchically \textit{increasing} the \textit{temporal receptive field} with \textit{depth}.
See \figref{teaser:b} for an illustration.

\subsection{Implementation Details}\label{sec:main:implement}
\paragraph{Data Loading.}
During both training and inference, we perform sequential reading of consecutive chunks of frames (clips) to process videos in an online fashion.
This is also the natural setting in a wide range of applications, \eg, robotics or recognition on live streaming video.
In our implementation, we simply concatenate all videos and read them sequentially.
In cases where the cached memory comes from the previous video (\ie, at the video boundary) we mask the memory to be zero.

\paragraph{Compression Module Design.}
The compression module can be any function that reduces the number of tokens but maintains the dimensionality $d$. 
In our instantiation we choose a learnable pooling~\cite{MViT} due to its simplicity and strong performance, but other choices are possible.
We will study the effect of different downsampling factors in \secref{ablation}.

\paragraph{Positional Embedding.}
In the original MViT~\cite{MViT}, absolute positional embeddings are added to the input of the network, 
and each clip uses the same positional embeddings.
Thus the positional embeddings can only indicate the positions within a clip, but not the order across multiple clips. 
We thus adopt the relative positional embedding used in ``the improved MViT"~\cite{li2021improved}, 
so that memory at different points in time has a different relative distance to the queries.

	\begin{table*}[t]
		\small
		\subfloat[\textbf{Per-layer memory length}\label{tab:abl:len}]{%
			\tablestyle{1.0pt}{1.05}
			\begin{tabular}{@{}
					lx{50}x{40}x{40}
					@{}}
				Mem len & Receptive field & GFLOPs & mAP\\
				\shline
				w/o mem & 1\x & 57.4 & 27.0\\
				1 & 8\x & 58.1 & 28.7\\
				\rowcolor{defaultcolor}
				2 & 16\x& 58.7 & \textbf{29.3}\\
				3 & 24\x & 59.3 & 29.2\\
				4 & 32\x & 60.0 & 28.8\\
				\\
				\\
			\end{tabular}
		}\hfill  %
		\subfloat[\textbf{Memory compression factor}\label{tab:abl:compression}]{%
			\tablestyle{1.0pt}{1.05}
			\begin{tabular}{@{}
					lx{40}x{40}
					@{}}
				Compress factor & GFLOPs  & mAP\\
				\shline
				none & 73.0 & 28.9\\
				1{\x}2{\x}2 & 62.3& 29.0\\
				2{\x}1{\x}1 & 65.3& 29.1\\
				2{\x}2{\x}2 & 59.9& 29.0\\
				2{\x}4{\x}4 & 58.2& 28.3\\
				\rowcolor{defaultcolor}
				4{\x}2{\x}2 & 58.7 & \textbf{29.3}\\
				4{\x}4{\x}4 & 57.8& 28.6\\
			\end{tabular}
		}\hfill  %
		\subfloat[\textbf{Memory augmentation layers}\label{tab:abl:location}]{%
			\tablestyle{1.0pt}{1.05}
			\begin{tabular}{@{}
					lx{40}x{40}
					@{}}
				Aug layers & GFLOPs & mAP\\
				\shline
				all & 60.2 & 29.1\\
				75\% (uniform) & 59.5& 29.1\\
				\rowcolor{defaultcolor}
				50\% (uniform) & 58.7 & \textbf{29.3}\\
				25\% (uniform) & 58.1& 28.7\\
				early & 58.4 &  28.6\\
				middle & 58.8& 28.7\\
				late & 57.8& 29.1
			\end{tabular}
		}  %
		\caption{\textbf{Ablation Experiments.} We conduct detailed ablation on (a): per-layer memory length, (b): compression module downsampling factors, and (c): layers to augment memory. All results are on conducted on the AVA dataset~\cite{Gu2018} with Kinetics-400~\cite{Kay2017} pre-training. 
		We see that MeMViT can increase receptive field, and thus performance, clearly with only small computational cost on a wide range of different design choices. The \hl{gray rows} denote default choices. (mAP in \%).
		\vspace{0mm}}
		\label{tab:abl}

	\end{table*}

\section{Experiments}
\label{sec:exp}
In this section, we will first compare the scaling behavior of MeMViT with other strategies in \secref{strategies}
and then ablate different design choices of MeMViT in \secref{ablation}.
We perform these experiments on the AVA spatiotemporal action localization dataset~\cite{Gu2018}, which consists of 299 15-minute-long videos sampled from movies.
In \secref{generalization}, we will study how our method, developed on AVA, 
generalizes on multiple other tasks and datasets.
We will finally compare MeMViT to prior state-of-the-art methods in \secref{stoa}.

\paragraph{Implementations.}
Our default MeMViT model is based on MViT-B~\cite{MViT,li2021improved} (16 layers) with 16-frame input clips, sampled at a temporal stride of 4 (denoted `16\x4' in model specifications).
We follow improvements proposed in Li~\etal~\cite{li2021improved} due to stronger performance.
Following prior work~\cite{Wu2019,feichtenhofer2020x3d,Feichtenhofer2019,MViT}, all models in this section are pre-trained on Kinetics-400~\cite{Kay2017} unless otherwise stated.
The AVA models are trained for 30 epochs with SGD using a batch size of 128.
We apply random horizontal flipping and random cropping of size 224$^2$ from frames resized such that the short side $\in[256, 340]$ as data augmentation.
We report FLOPs on 224$^2$ crops.
We use a cosine learning rate schedule with a base learning rate of 0.6 and weight decay of 10$^{-8}$.
All runtime and memory usages are measured on the same machine with an NVIDIA 16-GB Quadro GP100 GPU with batch size of one.
The Kinetics pre-training details, AVA person detector specifications, and additional details are available in the Appendix.
All methods are implemented using PySlowFast~\cite{fan2020pyslowfast}.

\subsection{Scaling Strategies}\label{sec:strategies}
We first compare the scaling behavior of MeMViT with the widely used ``baseline scaling" method~\cite{Wang2018,Feichtenhofer2019}, which increases the temporal support of a video model by increasing the number of frames $T$ in its input.
In \figref{tradeoff}, we see that by increasing $M$, MeMViT scales up to significantly longer temporal support with greatly lower training GPU memory (\figref{tradeoff:trmem}), inference GPU memory (\ref{fig:tradeoff:temem}), training runtime (\ref{fig:tradeoff:trtime}), inference runtime (\ref{fig:tradeoff:tetime}) and FLOPs (\ref{fig:tradeoff:flops}).
\figref{tradeoff:map} shows that under the same computational costs, our method also obtains clearly \emph{better accuracy}.
We also see that our compression method brings a clear trade-off improvement over the ``basic" version that does not compress memory.
These results demonstrate that our memory-based design with compression is a promising direction to build practical and strong long-term video models.

\subsection{Ablation Experiments}\label{sec:ablation}
\paragraph{Per-Layer Memory Length.}
\tabref{abl:len} compares models with different per-layer memory length ($M$).
We see that all models augmented with memory enjoy clear improvement over the baseline short-term model (1.7-2.3\% absolute gain in mAP).
Interestingly, the behavior is not very sensitive to the choice of the memory length.
Using a per-layer memory length of 2, which corresponds to 16\x~larger (36-second) receptive field, results in best performance for AVA\@.
We use $M{=}$2 as default in the following AVA experiments.

\paragraph{Memory Compression Factor.}
\tabref{abl:compression} compares compression modules with different downsampling factors.
We see that temporal downsampling can be slightly more aggressive (4\x) than spatial downsampling (2\x) while achieving strong performance. 
Interestingly, our compression method actually \emph{improves} the accuracy over the model without compression.
This supports our hypothesis that learning `what to keep' in memory can potentially suppress irrelevant noise and help learning.
We use downsampling factor of 4\x2\x2 (for time, height, and width, respectively) as default due to its strong performance.

\paragraph{Memory Augmentation Layers.}
In \tabref{abl:location}, we explore if we need to augment memory at all attention layers, and if not, adding memory at which layers is most effective. 
Interestingly, we see that attending memory at all layers is unnecessary.\footnote{Interestingly, similar findings are seen in NLP literatures in the context of language modeling~\cite{rae2020transformers}.}
In fact, augmenting 50\% of the layers (\ie, alternating between normal self- and memory-augmented attention) leads to the best performance while saving computation.
Furthermore, we observe that putting them uniformly throughout the network works slightly better than concentrating them at early (stage 1\&2) layers, middle (stage 3) layers, or late (stage 4) layers.

	\begin{figure}[t]
		\resizebox{1\linewidth}{!}{
			\hspace{4mm}
			\subfloat[\textbf{}\label{fig:pipeline:mem}]{%
				\begin{tikzpicture}
				\begin{axis}[
				clip=false,
				width=0.24\textwidth,
				height=0.24\textwidth,
				xlabel={Temporal support (s)},
				ylabel={Train GPU mem (GB)},
				xmin=0,
				ymin=0,
				every axis plot/.append style={very thick,mark options={scale=0.5, solid}},
				legend style={at={(1.2,0.56)},anchor=north east, cells={align=left}}
				]
				\addplot+[skyblue2,mark=pentagon*,mark options={solid}, solid]
				table[row sep=crcr] {
					x y\\
					2.0	4.15\\
70.4	4.96\\
138.7	5.73\\
206.9	6.5\\
275.2	7.28\\
				};

				\addplot+[chameleon3,mark=diamond*,mark options={solid, fill=chameleon3}, densely dashdotted]
				table[row sep=crcr] {
					x y\\
					2.0	4.15\\
70.4	4.66\\
138.7	5.05\\
206.9	5.45\\
275.2	5.86\\
				};
				
				\legend{w/o pipeline, \textbf{w/ pipeline}\\\textbf{(default)}}
				
				\end{axis}
				\end{tikzpicture}
			}\hspace{4mm}
			\subfloat[\textbf{}\label{fig:pipeline:time}]{%
				\begin{tikzpicture}
				\begin{axis}[
				clip=false,
				width=0.24\textwidth,
				height=0.24\textwidth,
				xlabel={Temporal support (s)},
				ylabel={Train iter time (s)},
				every axis plot/.append style={very thick,mark options={scale=0.5, solid}},
				xmin=0,
				ymin=0,
				cycle multiindex* list={%
					mycolormarklist
					\nextlist
					mystylelist
				}
				]
				\addplot+[skyblue2,mark=pentagon*,mark options={solid}, solid]
				table[row sep=crcr] {
					x y\\
					2.0	0.92\\
70.4	1.75\\
138.7	2.63\\
206.9	3.62\\
275.2	4.67\\
				};

				\addplot+[chameleon3,mark=diamond*,mark options={solid, fill=chameleon3}, densely dashdotted]
				table[row sep=crcr] {
					x y\\
					2.0	0.92\\
70.4	1.33\\
138.7	1.66\\
206.9	2.08\\
275.2	2.61\\
				};
				
				\end{axis}
				\end{tikzpicture}
			}
			\hspace{4mm}
		}\vspace{-3mm}
		\caption{\textbf{Compression Strategy.} Even with our relatively lightweight pooling-based compression module, 
the pipelined strategy already shows a significantly better scaling behavior in terms of both GPU memory usage (\figref{pipeline:mem}) and runtime (\figref{pipeline:time}).}\label{fig:pipeline}
\vspace{-3mm}
	\end{figure}
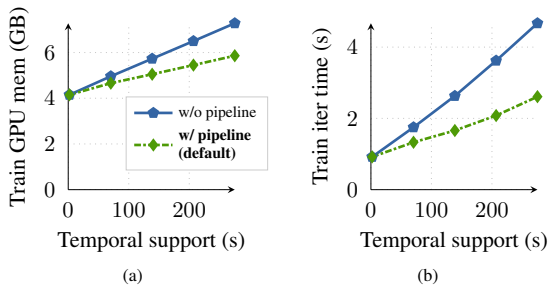

\begin{table*}[t]
\small
\resizebox{1.0\linewidth}{!}{
\hspace{-2.5mm}
\subfloat[\textbf{Additional pre-training datasets and model sizes.}\label{tab:gen:scale}]{%
	\tablestyle{3.0pt}{1.05}
\begin{tabular}{@{}y{20}lx{23}x{23}x{23}@{}}
Pre-train & Model & mAP (\%)  & GFLOPs & Param (M) \\ 
	\shline
	\multirow{2}{*}{K400} & MViT-16, 16\x4 & 27.0 & 57.4 & 34.5\\		
	 &  \textbf{MeMViT}-16, 16\x4 & \textbf{29.3} & 58.7 & 35.4\\
	\hline
	\multirow{2}{*}{K600} & MViT-24, 32\x3 & 30.1 & 204.4 & 51.3\\
	& \textbf{MeMViT}-24, 32\x3 & \textbf{32.3} &211.7 &52.6\\
	\hline
	\multirow{2}{*}{K700} & MViT-24, 32\x3 & 32.5 & 204.4 & 51.3\\
	& \textbf{MeMViT}-24, 32\x3 & \textbf{34.4} &211.7 &52.6\\
\end{tabular}
}
\hspace{2mm}
		\subfloat[\textbf{Additional Datasets \& Tasks}\label{tab:gen:data}]{%
			\tablestyle{1.0pt}{1.05}
\begin{tabular}{@{}llx{40}x{40}x{40}x{40}x{40}x{40}@{}}
			Task & Model & Action & Verb & Noun & Tail action & Tail verb & Tail noun\\ 
			\shline
			\multirow{2}{*}{AVA Loc.} & MViT & 27.0\dummy & - & - & - & - & -\\
			& \textbf{MeMViT} & \textbf{29.3} \gain{\textbf{2.3}} & - & -& -& -& -\\
			\hline
			\multirow{2}{*}{EPIC Cls.}& MViT & 44.6\dummy & 69.7\dummy & 56.1\dummy & - & - & -\\
			&\textbf{MeMViT} &\textbf{46.2} \gain{1.6} & \textbf{70.6} \gain{0.9} & \textbf{58.5}
			\gain{\textbf{2.4}}& -& -& -\\
			\hline
			\multirow{2}{*}{EPIC Anticip.\ \ } & MViT & 14.6\dummy & 29.3\dummy & 31.8\dummy & 12.2\dummy & 22.6\dummy & 25.5\dummy\\
			&\textbf{MeMViT} &\textbf{15.1} \gain{0.5} & \textbf{32.8} \gain{\textbf{3.5}} & \textbf{33.2} \gain{1.4} &
			\textbf{13.2} \gain{1.0} &
			\textbf{26.3} \gain{\textbf{3.7}}&
			\textbf{27.4} \gain{1.9}
			\\
			\end{tabular}}
		}\vspace{-2mm}  %
		\caption{\textbf{Generalization Analysis.}
		We show that our method
		brings consistent gains with different model sizes and pre-training datasets in \tabref{gen:scale}, and datasets and tasks in \tabref{gen:data}.
		Performance measured by mAP (\%) for AVA, top-1 (\%) for EPIC-Kitchens Classification, and class-mean recall@5 (\%)~\cite{furnari2018leveraging} for EPIC-Kitchens Anticipation following standard practice.\vspace{-2mm}}
		\label{tab:gen}
	\end{table*}

\paragraph{Compression Strategy.}
Finally, we compare the scaling behavior of our pipelined compression strategy with that of the basic version without pipeline in \figref{pipeline}.
We can see that even with our relatively lightweight pooling-based compression module, 
the pipelined strategy already shows a significantly better scaling behavior in terms of both GPU memory usage (\figref{pipeline:mem}) and runtime (\figref{pipeline:time}).
We thus use it by default in MeMViT.
We hope the better scaling behavior will help future research to scale up to even longer-term video models or explore more advanced compression modules more easily.

	\begin{table}[h!]
		\centering
		\tablestyle{2.5pt}{1.05}
		\begin{tabular}{@{}l|x{25}|x{21}x{21}x{21}x{21}@{}}
			Model & Pre- &   \multicolumn{2}{c}{mAP (\%)}  & FLOPs & Param \\ 
			 & train & center & full  & (G) & (M) \\ 
			\shline
			{SlowFast}, 4\x16, R50 \cite{Feichtenhofer2019} & \multirow{9}{*}{K400} & 21.9 &-& 52.6 & 33.7 \\
			SlowFast, 8\x8, R50~\cite{Feichtenhofer2019} &  & 22.7 &-& 96.9 & 33.8 \\ 
			SlowFast, 8\x8, R101~\cite{Feichtenhofer2019}   &  & 23.8  &-& 137.7 & 53.0 \\ 
			WOO, SFR50~\cite{chen2021watch} & & 25.4 &-&147.5 & -\\
			MViTv1-B, 16\x4~\cite{MViT}   &  & 24.5 & -&70.5 & 36.4 \\ 
			MViTv1-B, 32\x3~\cite{MViT}    &   & 26.8 & -&169.8 & 36.4 \\ 
			MViTv1-B, 64\x3~\cite{MViT}   &   & {27.3} & -&454.7 & 36.4 \\
			MViT-16, 16\x4~\cite{li2021improved} & & 26.2&27.0 &57.4 & 34.5\\
			\textbf{MeMViT}-16, 16\x4 & & \textbf{28.5}&\textbf{29.3} & 58.7 & 35.4\\		
			\hline
			
			{SlowFast}, 8\x8 R101+NL \cite{Feichtenhofer2019}   & \multirow{12}{*}{K600}   &  27.1 &-& 146.6 & 59.2 \\
			{SlowFast}, 16\x8 R101+NL \cite{Feichtenhofer2019}   &    &  27.5 &-& 296.3 & 59.2 \\
			X3D-XL~\cite{feichtenhofer2020x3d}  &    &  27.4 &-& 48.4 & 11.0 \\
			WOO, SFR101~\cite{chen2021watch} & & 28.3 &-& 251.7 & - \\
			{MViTv1}-B, 16\x4~\cite{MViT}   &  & 26.1 & -&70.4 & 36.3 \\ 
			{MViTv1}-B, 32\x3~\cite{MViT}    &   & 27.5 & -&169.8 & 36.4 \\  
			{MViTv1}-B-24, 32\x3~\cite{MViT}    &   & 28.7 &-& 236.0 & 52.9 \\  
			Object Transformer~\cite{wu2021towards} & & 31.0 &-& 243.8 & 86.2\\
			ACAR 8\x8, R101-NL~\cite{pan2021actor} & &-& 31.4 & 293.2$^\dagger$ & 118.4$^\dagger$\\
			MViT-24, 32\x3~\cite{li2021improved} & &29.4& 30.1 & 204.4 & 51.3\\		
			\textbf{MeMViT}-24, 32\x3 &  &31.5& {32.3} & 211.7 & 52.6\\		
			\textbf{MeMViT}-24, 32\x3, $\uparrow$312$^2$ &   &\textbf{32.8}& \textbf{33.6} & 620.0 & 52.6\\		

			\hline
			AIA~\cite{tang2020asynchronous} &\multirow{5}{*}{K700}& 32.3 &-& - & -\\
			ACAR R101~\cite{pan2021actor} &&-& 33.3 & 212.0$^\dagger$ & 107.4$^\dagger$\\
			MViT-24, 32\x3~\cite{li2021improved} & &31.8& 32.5 & 204.4 & 51.3\\		
			\textbf{MeMViT}-24, 32\x3 &&33.5 &{34.4} & {211.7} & {52.6}\\		
			\textbf{MeMViT}-24, 32\x3, $\uparrow$312$^2$ & & \textbf{34.4}&\textbf{35.4} & 620.0 & 52.6
		\end{tabular}
		\vspace{-3mm}
		\caption{\textbf{Comparison to prior work on AVA v2.2~\cite{Gu2018}}. 
		$^\dagger$: \footnotesize{ACAR does not provide parameters and flops but we estimate a lower bound calculating their `backbone' only, which contains two ``8\x8 R101-NL" (or ``8\x8 R101") SlowFast backbones for K600- (or K700-) pretraining.}
		\vspace{-4mm}}
		\label{tab:ava}
	\end{table}
	\begin{table*}[t]
		\centering
		\vspace{-5pt}
		\tablestyle{0.5pt}{1.05}
		\begin{tabular}{@{\extracolsep{4pt}}lx{82}x{22}x{22}x{22}x{22}x{22}x{22}x{22}x{22}x{22}x{22}@{}}
			Model & External data / & Param & \multicolumn{3}{c}{Overall}& \multicolumn{3}{c}{Unseen}& \multicolumn{3}{c}{Tail} \\
			\cline{4-6}\cline{7-9}\cline{10-12} 
			& extra annotations & (M) & Action &   Verb  & Noun& Action &   Verb  & Noun& Action &   Verb  & Noun \\ 
			\shline
			\demph{TempAgg (RGB + Obj + Flow + ROI)}~\cite{sener2021technical} & \demph{IN1K + EPIC boxes}& \demph{-}&\demph{14.7}&\demph{23.2}&\demph{31.4}&\demph{14.5}&\demph{28.0}&\demph{26.2}&\demph{11.8}&\demph{14.5}&\demph{22.5} \\
			\demph{RULSTM (RGB + Obj + Flow)~\cite{furnari2020rolling}}\quad\quad  & \demph{IN1K + EPIC boxes} & \demph{-}& \demph{14.0}&\demph{27.8} &\demph{30.8}&\demph{14.2} & \demph{28.8} & \demph{27.2} & \demph{11.1} & \demph{19.8} & \demph{22.0} \\
			\demph{TSN-AVT+ (RGB + Obj)~\cite{girdhar2021anticipative}} & \demph{IN21K + EPIC boxes} & \demph{-}& \demph{14.8}&\demph{25.5}&\demph{31.8}	&\demph{11.5} & \demph{25.5} & \demph{23.6} & \demph{12.6} & \demph{18.5} & \demph{25.8} \\
			\demph{AVT+ (RGB + Obj)~\cite{girdhar2021anticipative}}&  \demph{IN21K + EPIC boxes}& \demph{-} &\demph{15.9} & \demph{28.2} &\demph{32.0} & \demph{11.9} & \demph{29.5} & \demph{23.9} & \demph{14.1} & \demph{21.1} & \demph{25.8} \\
			\hline
			chance & -&-&0.2 & 6.4 & 2.0 & 0.5 & 14.4 & 2.9 & 0.1 & 1.6 & 0.2\\
			TempAgg (RGB)~\cite{sener2021technical} & IN1K&-&13.0&24.2&29.8&12.2&27.0&23.0&10.4&16.2&22.9\\
			AVT (RGB)~\cite{girdhar2021anticipative} & IN21K& 378 &  14.9 &30.2 & 31.7 & -  & -  & -  & -  & -  & - \\
			\textbf{MeMViT}, 16\x4 & K400 & \textbf{59}& 15.1 & \textbf{32.8} & 33.2 &  9.8 &27.5 &21.7 &13.2 &\textbf{26.3} &27.4 \\
			\textbf{MeMViT}, 32\x3 & K700 & 212 &\textbf{17.7} &32.2 &\textbf{37.0} &\textbf{15.2} &\textbf{28.6} &\textbf{27.4} &\textbf{15.5} &25.3 &\textbf{31.0} \\
		\end{tabular}
		\vspace{-2mm}
		\caption{\textbf{Comparison to prior work on EPIC-Kitchens-100 Action Anticipation~\cite{Damen2018EPICKITCHENS,Damen2021PAMI}}. 
		Accuracy measured by class-mean recall@5 (\%)~\cite{furnari2018leveraging} following the standard protocol~\cite{Damen2021PAMI}.
		\demph{Gray} denotes challenge entries that use additional modalities, such as optical flow or separately extracted object features;
		MeMViT uses only pixels and still outperforms all of them.}
		\label{tab:anti}
		\vspace{-2mm}
	\end{table*}

\subsection{Generalization Analysis}\label{sec:generalization}
So far, we developed and analyzed our method mainly based on an MViT-B~\cite{MViT} default backbone on the AVA action localization dataset~\cite{Gu2018}.
Next, we examine MeMViT's ability to generalize to different settings.

\paragraph{Additional Pre-training Datasets and Model Sizes.}
We first examine how our method generalizes to different pre-training datasets and model sizes.
In particular, we grow both our pre-training dataset from the K400 dataset~\cite{Kay2017} (400 classes; $\app$240k videos) to the 
K600 dataset~\cite{Carreira2018} (600 classes; $\app$387k videos) and the 
K700 dataset~\cite{Carreira19} (700 classes; $\app$522k videos), 
and also our base model from 16 layers with 16\x4 inputs (denoted `MeMViT-16, 16\x4') to 24 layers with 32\x3 inputs (denoted `MeMViT-24, 32\x3').
Training recipe stays the same.
See the Appendix for detailed model specification for MeMViT-24.
\tabref{gen:scale} shows that despite the different settings,
MeMViT provides consistent performance gain over the original short-term model (MViT), 
suggesting good generalizability of our method. 

\paragraph{Additional Datasets and Tasks.}
\tabref{gen:data} presents results on EPIC-Kitchens-100 egocentric action classification and EPIC-Kitchens-100 action anticipation~\cite{Damen2018EPICKITCHENS,Damen2021PAMI}. 
The models used here are the same ``MeMViT-16, 16\x4" as the default model used for AVA\@, except that for EPIC-Kitchens we found that a longer-term model that uses $M=4$ (32\x longer-term, or 70.4-second receptive field) to work the best.
The model for action anticipation is a causal version to make sure the model output does not see frames beyond the ``observed video"~\cite{Damen2021PAMI}.
Complete model and training details are available in the Appendix.
Note that recognition on \emph{egocentric} videos in the EPIC-Kitchens dataset is quite challenging due to severe motion blur and occlusions on the target action~\cite{Damen2018EPICKITCHENS,Damen2021PAMI}. 
Also note the large domain difference compared to the videos in AVA~\cite{Gu2018}, which contains stable movie content with different camera motion.

Despite the differences,
we see that MeMViT, developed on AVA, works well out-of-the-box on EPIC-Kitchens as well.
If we take a closer look at the EPIC \emph{classification} task, we see that `noun' recognition is a harder task than `verb' recognition, potentially because objects are often occluded by hands, blurred, or even out of the scene.
Nonetheless, MeMViT boosts `noun' recognition significantly (\textbf{+2.4}\%), supporting our hypothesis that MeMViT may utilize long-term context to disambiguate objects.
On the other hand, for action \emph{anticipation}, 
predicting the verbs is actually more challenging than predicting the nouns, potentially because nouns are more persistent but verbs can change more frequently (consider `washing tomatoes', followed by `cutting tomatoes', followed by `putting tomatoes (into something)'). 
While with a short-term model, predicting the next `verb' given \emph{only} the previous one might be challenging, 
MeMViT sees much more context into the past, bringing large improvement on verbs (\textbf{+3.5}\%) and tail verbs (\textbf{+3.7}\%).

\subsection{State-of-the-Art Comparison}\label{sec:stoa}

\paragraph{The AVA Dataset.}
\tabref{ava} compares MeMViT with prior work on the AVA v2.2 dataset~\cite{Gu2018}.
We see that under all pre-training settings,
MeMViT obtains a significantly higher accuracy than prior work while having a comparable or lower number of FLOPs and parameters.
In particular, it outperforms ACAR~\cite{pan2021actor} ---the state-of-the-art `long-term feature-bank'-based approach---
without requiring two backbones, additional feature-bank model training, and additional feature bank extraction.
If we further fine-tune MeMViT (trained on 224$^2$ crops) on higher resolution of 312$^2$,
the single model achieves \textbf{35.4} mAP.

	\begin{table}[t]
		\centering
		\resizebox{1\linewidth}{!}{
		\hspace{-2.5mm}
		\tablestyle{4pt}{1.05}
		\begin{tabular}{@{}lx{35}x{13}x{13}x{13}x{21}x{12}x{16}x{16}@{}}
			Model & Pre-train &Act. &   Verb  & Noun & Run-time(s) & Mem (GB) & FLOPs (G) & Param (M) \\ 
			\shline
			TSN~\cite{Wang2016a} &IN1K& 33.2 &60.2 &46.0& - &-&-&-\\
			TempAgg~\cite{sener2021technical} &IN1K & 36.9& 59.9 &45.1& -&-&-&-\\
			TSM~\cite{lin2019tsm} &IN1K&38.3 &67.9 &49.0& - &-&-&-\\
			SlowFast~\cite{Feichtenhofer2019}& K400 & 38.5& 65.6 &50.0 &-& -&-&-\\
			Ego-Exo~\cite{ego-exo}      & K400 & - & 67.0 & 52.9 &-& -&-&- \\
			IPL~\cite{wang2021interactive} & K400&41.0 &68.6 &51.2 & - &- &- &-\\
			ViViT-L/16\x2~\cite{arnab2021vivit} &IN21K&44.0 &66.4 &56.8 & -&-&3410 &100\\
			MFormer~\cite{MotionFormer} & \scriptsize{IN21K+K400} & 43.1 & 66.7 & 56.5 & - & -&370& 109\\
			MFormer-HR~\cite{MotionFormer} & \scriptsize{IN21K+K400} & 44.5 & 67.0 & \textbf{58.5} & - & -&959& 382\\
			MoViNet-A5~\cite{kondratyuk2021movinets} &N/A& 44.5 & 69.1 & 55.1 &0.49 &8.3&74.9 & {15.7}\\
			\textbf{MeMViT}, 16\x4 &K400&\textbf{46.2} & \textbf{70.6} & \textbf{58.5} & \textbf{0.16} &\textbf{1.7}&58.7 & 35.4\\
			\hline
			MoViNet-A6~\cite{kondratyuk2021movinets} &N/A& 47.7 & \textbf{72.2} &57.3 & 0.85 &8.3&117.0 &31.4\\
			\textbf{MeMViT}, 32\x3 &K600& \textbf{48.4} & {71.4}	&\textbf{60.3} &\textbf{0.35}&\textbf{3.9}& 211.7 & 52.6\\
		\end{tabular}}
		\caption{\textbf{Comparison to prior work on EPIC-Kitchens-100 Action Classification~\cite{Damen2018EPICKITCHENS,Damen2021PAMI}}. Accuracy measured by top-1 classification accuracy (\%). }
		\label{tab:epiccls}
	\end{table}
	
\paragraph{The EPIC-Kitchens-100 Action Classification Task.}
We next compare with prior work on EPIC-Kitchens-100 classification~\cite{Damen2018EPICKITCHENS,Damen2021PAMI}.
\tabref{epiccls} shows that MeMViT again outperforms all prior works, including both CNN-based~\cite{Wang2016a,Feichtenhofer2019,kondratyuk2021movinets,ego-exo} and ViT-based methods~\cite{arnab2021vivit,MotionFormer}.
In particular, the previous best method, MoViNet~\cite{kondratyuk2021movinets}, also considers an `online'-style model but using causal convolutions,
which extend the context only by half of the kernel size (typically one pixel) per layer, thus having a significantly shorter temporal support.
MeMViT works significantly better.
Also note that MoViNets' low FLOPs does not translate to efficient runtime on GPUs, in part because MoViNet extensively uses depthwise convolutions, which are known to have low FLOPs, but high runtime in practice~\cite{ilija_2020}.
MeMViT outperforms MoViNet by a clear margin while being 3\x~faster and at 2-5\x lower GPU memory.

While obtaining high performance, we emphasize that MeMViT uses a simpler and lighter testing procedure, where it simply perform \emph{one pass}
of the videos sequentially, and aggregate all predictions made on target segments by average pooling, without multi-crop testing or over-sampling on testing segments.

\paragraph{The EPIC-Kitchens-100 Action Anticipation Task.}
Finally, we compare MeMViT with prior work on EPIC-Kitchens-100 Anticipation~\cite{Damen2018EPICKITCHENS,Damen2021PAMI}.
Here we use our default model (MeMViT-16, 16\x4) pre-trained on Kinetics-400~\cite{Kay2017} and also a larger
MeMViT-24, 32\x3, pre-trained on Kinetics-700~\cite{Carreira19}.
\tabref{anti} shows that MeMViT outperforms all prior work, including those that use multiple modalities, such as optical flow~\cite{furnari2020rolling}, separately trained object feature extractors~\cite{girdhar2021anticipative} and large-scale pre-training (IN-21K~\cite{deng2009imagenet} has $\app$60\x~more labels than K400).

The competition winner this year, AVT+~\cite{girdhar2021anticipative}, uses a large ViT-based backbone with IN21K pre-training that additionally uses auxiliary losses (\eg, feature regression loss and action recognition loss) and object features. 
With a simple cross-entropy loss on action labels, our long-term MeMViT outperforms AVT+ by a large margin (action: \textbf{+1.8}\%, verb: \textbf{+4.0}\%, noun: \textbf{+5.0}\%).

\section{Conclusion}
Long-term video understanding is an important goal for computer vision.
To get there, having a practical model for long-term visual modeling is a basic prerequisite.
In this paper, we show that extending existing state-of-the-art models to include more input frames does not scale well. 
Our memory-based approach, MeMViT, scales much more efficiently and achieves better accuracy.
The techniques presented in this paper are general and applicable to other transformer-based video models.
We hope MeMViT will be useful for future long-term video modeling research.

\appendix
\section{Appendix}

\renewcommand\thefigure{\thesection.\arabic{figure}}
\renewcommand\thetable{\thesection.\arabic{table}}
\setcounter{figure}{0} 
\setcounter{table}{0} 

\setcounter{table}{0}
\renewcommand{\thetable}{A.\arabic{table}}

\subsection{Architecture Specifications}
The architecture design of MeMViT is based on MViT~\cite{MViT} with improvements proposed in Li~\etal~\cite{li2021improved}.
\tabref{archi} presents the exact specification.
\begin{table}[h!]
	\centering
	\subfloat[\textbf{MeMViT-16, 16\x4}]{%
		\tablestyle{0.5pt}{1.05}
		\tablestyle{1pt}{1.08}  \scriptsize 
		\begin{tabular}{c|c|c}
			stage & operators & output sizes \\
			\shline
			\multirow{1}{*}{data} & \multirow{1}{*}{stride \tcolor{4}\x1\x1}   &  \outsizesRaw{\tcolor{\textbf{16}}}{\xycolor{224}}{\xycolor{224}}{1}   \\
			\hline
			
			\multirow{2}{*}{cube$_1$} & \multicolumn{1}{c|}{3\x7\x7, {96}} &    \outsizesRawD{\wcolor{$96$}}{\tcolor{\textbf{8}}}{\xycolor{56}}{\xycolor{56}}{2}    \\
			& stride 2\x4\x4   \\
			\hline
			\multirow{2}{*}{scale$_2$}  & \blockatta{96}{{384}}{\textbf{1}} & \outsizesRawD{\wcolor{$96$}}{\tcolor{\textbf{8}}}{\xycolor{56}}{\xycolor{56}}{2}  \\
			&  & \\
			\hline
			\multirow{2}{*}{scale$_3$}  & \blockatta{192}{{768}}{\textbf{2}} & \outsizesRawD{\wcolor{$192$}}{\tcolor{\textbf{8}}}{\xycolor{28}}{\xycolor{28}}{2}  \\
			&  & \\
			\hline
			\multirow{2}{*}{scale$_4$}  & \blockatta{384}{{1536}}{\textbf{11}} & \outsizesRawD{\wcolor{$384$}}{\tcolor{\textbf{8}}}{\xycolor{14}}{\xycolor{14}}{2}  \\
			&  & \\
			\hline
			\multirow{2}{*}{scale$_5$}  & \blockatta{768}{{3072}}{\textbf{2}} & \outsizesRawD{\wcolor{$768$}}{\tcolor{\textbf{8}}}{\xycolor{7}}{\xycolor{7}}{2}  \\
			&  & \\
			\hline
		\end{tabular}
		}%
\vspace{2mm}
\subfloat[\textbf{MeMViT-24, 32\x3}]{%

		\tablestyle{0.5pt}{1.05}
		\tablestyle{1pt}{1.08}  \scriptsize 
		\begin{tabular}{c|c|c}
			stage & operators & output sizes \\
			\shline
			\multirow{1}{*}{data} & \multirow{1}{*}{stride \tcolor{4}\x1\x1}   &  \outsizesRaw{\tcolor{\textbf{32}}}{\xycolor{224}}{\xycolor{224}}{1}   \\
			\hline
			
			\multirow{2}{*}{cube$_1$} & \multicolumn{1}{c|}{3\x7\x7, {96}} &    \outsizesRawD{\wcolor{$96$}}{\tcolor{\textbf{16}}}{\xycolor{56}}{\xycolor{56}}{2}    \\
			& stride 2\x4\x4   \\
			\hline
			\multirow{2}{*}{scale$_2$}  & \blockatta{96}{{384}}{\textbf{2}} & \outsizesRawD{\wcolor{$96$}}{\tcolor{\textbf{16}}}{\xycolor{56}}{\xycolor{56}}{2}  \\
			&  & \\
			\hline
			\multirow{2}{*}{scale$_3$}  & \blockatta{192}{{768}}{\textbf{3}} & \outsizesRawD{\wcolor{$192$}}{\tcolor{\textbf{16}}}{\xycolor{28}}{\xycolor{28}}{2}  \\
			&  & \\
			\hline
			\multirow{2}{*}{scale$_4$}  & \blockatta{384}{{1536}}{\textbf{16}} & \outsizesRawD{\wcolor{$384$}}{\tcolor{\textbf{16}}}{\xycolor{14}}{\xycolor{14}}{2}  \\
			&  & \\
			\hline
			\multirow{2}{*}{scale$_5$}  & \blockatta{768}{{3072}}{\textbf{3}} & \outsizesRawD{\wcolor{$768$}}{\tcolor{\textbf{16}}}{\xycolor{7}}{\xycolor{7}}{2}  \\
			&  & \\
			\hline
		\end{tabular}
		}%
		\caption{\textbf{Architecture specification} for our ``MeMViT-16, 16\x4" (default) and ``MeMViT-24, 32\x3" models. Bold face highlights the difference between the two (\ie, temporal resolution and depth). MHPA($c$): Multi-Head Pooling Attention~\cite{MViT} with $c$ channels. MLP($c'$): MultiLayer Perceptron with $c'$ channels.}
		\label{tab:archi}
\end{table}

\paragraph{Relative Positional Embeddings.}
As discussed in \secref{main:implement}, 
we use relative positional embeddings instead of absolute positional embeddings as used in MViT~\cite{MViT}.
Our implementation is based on Shaw~\etal~\cite{shaw-etal-2018-self}, \ie,\footnote{The only difference between our implementation and Shaw~\etal~\cite{shaw-etal-2018-self} is that we do not add the additional embeddings on ``values", as in preliminary experiments we did not find it to improve accuracy.}
\begin{align*}
\mathrm{Attn}(Q, K, V) = \mathrm{Softmax}\rbr{(QK^\top + E^{(\mathrm{rel})}) / \sqrt{d}}V,
\end{align*}
\vspace{-12pt}
\begin{align}
\text{where \quad} E^{(\mathrm{rel})}_{ij} = Q_i \cdot R_{p(i),p(j)}.
\end{align}
$p(i)$ and $p(j)$ denote the spatiotemporal positions of tokens $i$ (in queries) and $j$ (in keys/values), respectively.
In other words, we learn relative positional embeddings $R$ that interact with queries $Q$ depending on the relative positions between the queries and the keys/values.
Note, however, that the number of possible embeddings grows in $\mathcal{O}(T\times H\times W)$, which is significantly more expensive than the one-dimensional case considered in  Shaw~\etal~\cite{shaw-etal-2018-self} for language modeling.
We thus decompose the relative positional embeddings into
\begin{equation}\label{eq:rel}
{{R}}_{p(i), p(j)} = {{R}}_{t(i),t(j)}^{\mathrm{t}} + {{R}}_{h(i),h(j)}^{\mathrm{h}} + {{R}}_{w(i),w(j)}^{\mathrm{w}},
\end{equation}
where $R^t$, $R^h$, and $R^w$ denote the relative positional embeddings along the temporal, frame hight, and frame width dimensions, respectively.
$t(i)$, $h(i)$, $w(i)$ denote the temporal position, the vertical position, and the horizontal position of token $i$, respectively.

\subsection{Kinetics Pre-training Details}
To pre-train MeMViT on the Kinetics datasets~\cite{Kay2017,Carreira2018,Carreira19} efficiently,
we propose a progressive strategy.
Namely, instead of training on full Kinetics videos throughout, we progressively increase the video length from one clip long (randomly sampled from full video) to the full video (10 seconds for Kinetics).\footnote{When MeMViT operates on videos that are one-clip-long, it effectively falls back to a short-term MViT (since there is no memory about the video cached from the previous step).}
Intuitively, this strategy allows the model to see more diverse spatial patterns in earlier epochs for faster spatial pattern learning and gradually adapt to longer videos in later epochs.
Concretely, we extend the original MViT recipe (that trains on one-clip-long videos sampled from full videos) by a ``second stage", which contains 40 epochs with 4 epochs of warm-up~\cite{Goyal2017}. 
Within the 40 epochs, we train on videos that are 2-, 3-, 4-, and finally 5-clip-long for 10 epochs each.
For data augmentation, we randomly drop $m \in [0, M-1]$ steps out of the $M$ steps of memory tensors at each iteration of training.
(At inference time, we still use all $M$ steps of memory.)
All other optimization hyperparameters follow the original MViT recipe~\cite{MViT}.

\subsection{AVA Experiments}
\paragraph{Person Detector.}
The person detector used in AVA experiments is a Faster R-CNN~\cite{Ren2015} with a ResNeXt-101-FPN~\cite{Xie2017,Lin2017} backbone from Wu~\etal~\cite{Wu2019}.
The model obtains 93.9 AP@50 on the AVA validation set~\cite{Wu2019}.
Please refer to the original paper~\cite{Wu2019} for details.

\paragraph{Output Head.}
Instead of using a linear output head for AVA, we additionally add a transformer layer (namely, an MViT layer without pooling, since each token is already RoI-pooled) before the linear classifier.
We find this to improve accuracy. 
\tabref{abl} presents ablation results.

\subsection{EPIC-Kitchens-100 Experiments}
We train our EPIC-Kitchens models with AdamW~\cite{loshchilov2017decoupled} for 30 epochs using a base learning rate of 0.0002, a weight decay of 0.05, and a batch size of 128.
Other training hyperparameters follow the Kinetics~\cite{Kay2017} recipe of MViT~\cite{MViT}.
We fine-tune action anticipation models from action classification models using the same training recipe.

For the anticipation task,
we perform experiments on a \emph{causal} version of MeMViT, to make sure our prediction does not depend on frames beyond the ``observed video"~\cite{Damen2018EPICKITCHENS,Damen2021PAMI}.
In particular, we 1) modify the learnable pooling so that it strictly pools only current or past contents,
2) mask attention so that it attends only current or past contents, 3) make the convolutions in the data layer `causal', and 4) remove the global `classification token'.
Following common practice in the object detection community~\cite{tan20201st,tan2020equalization}, we use equalization loss~\cite{tan2020equalization} with threshold $\lambda=$ 0.003 to address the class imbalance issue.

Our action classification model has two heads to predict verb and noun, respectively, following prior work~\cite{Wu2019,arnab2021vivit}.
Our action anticipation model has only one head to predict the action directly
and marginalize the output probabilities to obtain the verb and noun predictions, following standard practice~\cite{furnari2020rolling,girdhar2021anticipative}.

\subsection{Supplementary Experiments}
\paragraph{Model Detail Ablation.}
\tabref{abl} presents additional ablation on our implementations choices.

\begin{table}[h!]
	\centering
	\tablestyle{4pt}{1.05}
	\begin{tabular}{@{}lx{35}@{}}
		 & mAP \\
		\shline
		MViT-B, 16\x4~\cite{MViT} &24.5\\
		\quad+ relative positional embedding & 25.4\\
		\quad+ pool first & 25.5\\
		\quad+ test on full frame & 26.6\\
		\quad+ attention head (\underline{our default baseline})\quad\quad\quad\quad\quad\quad& \textbf{27.0}\\
	\end{tabular}
	\vspace{-2mm}
	\caption{\textbf{Detailed ablation on our default baseline model}.}
	\label{tab:abl}
\end{table}

	{\small
		\bibliographystyle{ieee_fullname}
		\bibliography{memvit}

\begin{thebibliography}{10}\itemsep=-1pt

\bibitem{Abu-El-Haija2016}
Sami Abu-El-Haija, Nisarg Kothari, Joonseok Lee, Paul Natsev, George Toderici,
  Balakrishnan Varadarajan, and Sudheendra Vijayanarasimhan.
\newblock Youtube-8m: A large-scale video classification benchmark.
\newblock {\em arXiv:1609.08675}, 2016.

\bibitem{arnab2021vivit}
Anurag Arnab, Mostafa Dehghani, Georg Heigold, Chen Sun, Mario Lu{\v{c}}i{\'c},
  and Cordelia Schmid.
\newblock {ViViT}: A video vision transformer.
\newblock In {\em Proc. ICCV}, 2021.

\bibitem{bertasius2021space}
Gedas Bertasius, Heng Wang, and Lorenzo Torresani.
\newblock Is space-time attention all you need for video understanding?
\newblock In {\em Proc. ICCV}, 2021.

\bibitem{Carreira2018}
Joao Carreira, Eric Noland, Andras Banki-Horvath, Chloe Hillier, and Andrew
  Zisserman.
\newblock A short note about {Kinetics}-600.
\newblock {\em arXiv:1808.01340}, 2018.

\bibitem{Carreira19}
Jo{\~{a}}o Carreira, Eric Noland, Chloe Hillier, and Andrew Zisserman.
\newblock A short note on the kinetics-700 human action dataset.
\newblock {\em arXiv preprint arXiv:1907.06987}, 2019.

\bibitem{Carreira_2018_ECCV}
Joao Carreira, Viorica Patraucean, Laurent Mazare, Andrew Zisserman, and Simon
  Osindero.
\newblock Massively parallel video networks.
\newblock In {\em Proc. ECCV}, 2018.

\bibitem{chen2021watch}
Shoufa Chen, Peize Sun, Enze Xie, Chongjian Ge, Jiannan Wu, Lan Ma, Jiajun
  Shen, and Ping Luo.
\newblock Watch only once: An end-to-end video action detection framework.
\newblock In {\em Proc. ICCV}, 2021.

\bibitem{chen2020memory}
Yihong Chen, Yue Cao, Han Hu, and Liwei Wang.
\newblock Memory enhanced global-local aggregation for video object detection.
\newblock In {\em Proc. CVPR}, 2020.

\bibitem{chen2021visformer}
Zhengsu Chen, Lingxi Xie, Jianwei Niu, Xuefeng Liu, Longhui Wei, and Qi Tian.
\newblock Visformer: The vision-friendly transformer.
\newblock In {\em Proc. ICCV}, 2021.

\bibitem{cheng2019sparse}
Changmao Cheng, Chi Zhang, Yichen Wei, and Yu-Gang Jiang.
\newblock Sparse temporal causal convolution for efficient action modeling.
\newblock In {\em ACM MM}, 2019.

\bibitem{dai2019transformer}
Zihang Dai, Zhilin Yang, Yiming Yang, Jaime Carbonell, Quoc~V Le, and Ruslan
  Salakhutdinov.
\newblock Transformer-xl: Attentive language models beyond a fixed-length
  context.
\newblock In {\em ACL}, 2019.

\bibitem{Dalal2005}
Navneet Dalal and Bill Triggs.
\newblock Histograms of oriented gradients for human detection.
\newblock In {\em Proc. CVPR}, 2005.

\bibitem{Damen2018EPICKITCHENS}
Dima Damen, Hazel Doughty, Giovanni~Maria Farinella, Sanja Fidler, Antonino
  Furnari, Evangelos Kazakos, Davide Moltisanti, Jonathan Munro, Toby Perrett,
  Will Price, and Michael Wray.
\newblock Scaling egocentric vision: The epic-kitchens dataset.
\newblock In {\em ECCV}, 2018.

\bibitem{Damen2021PAMI}
Dima Damen, Hazel Doughty, Giovanni~Maria Farinella, Sanja Fidler, Antonino
  Furnari, Evangelos Kazakos, Davide Moltisanti, Jonathan Munro, Toby Perrett,
  Will Price, and Michael Wray.
\newblock The epic-kitchens dataset: Collection, challenges and baselines.
\newblock {\em PAMI}, 2021.

\bibitem{deng2009imagenet}
Jia Deng, Wei Dong, Richard Socher, Li-Jia Li, Kai Li, and Li Fei-Fei.
\newblock {ImageNet}: A large-scale hierarchical image database.
\newblock In {\em Proc. CVPR}, 2009.

\bibitem{dollar2005behavior}
Piotr Doll{\'a}r, Vincent Rabaud, Garrison Cottrell, and Serge Belongie.
\newblock Behavior recognition via sparse spatio-temporal features.
\newblock In {\em International Workshop on Visual Surveillance and Performance
  Evaluation of Tracking and Surveillance}, 2005.

\bibitem{Donahue2015}
Jeff Donahue, Lisa~Anne Hendricks, Sergio Guadarrama, Marcus Rohrbach,
  Subhashini Venugopalan, Kate Saenko, and Trevor Darrell.
\newblock Long-term recurrent convolutional networks for visual recognition and
  description.
\newblock In {\em Proc. CVPR}, 2015.

\bibitem{dong2021cswin}
Xiaoyi Dong, Jianmin Bao, Dongdong Chen, Weiming Zhang, Nenghai Yu, Lu Yuan,
  Dong Chen, and Baining Guo.
\newblock Cswin transformer: A general vision transformer backbone with
  cross-shaped windows.
\newblock {\em arXiv preprint arXiv:2107.00652}, 2021.

\bibitem{ViT}
Alexey Dosovitskiy, Lucas Beyer, Alexander Kolesnikov, Dirk Weissenborn,
  Xiaohua Zhai, Thomas Unterthiner, Mostafa Dehghani, Matthias Minderer, Georg
  Heigold, Sylvain Gelly, et~al.
\newblock An image is worth 16x16 words: Transformers for image recognition at
  scale.
\newblock In {\em Proc. ICLR}, 2021.

\bibitem{efros}
Alexei~A Efros, Alexander~C Berg, Greg Mori, and Jitendra Malik.
\newblock Recognizing action at a distance.
\newblock In {\em Proc. ICCV}, 2003.

\bibitem{fan2020pyslowfast}
Haoqi Fan, Yanghao Li, Bo Xiong, Wan-Yen Lo, and Christoph Feichtenhofer.
\newblock {PySlowFast}.
\newblock \url{https://github.com/facebookresearch/slowfast}, 2020.

\bibitem{MViT}
Haoqi Fan, Bo Xiong, Karttikeya Mangalam, Yanghao Li, Zhicheng Yan, Jitendra
  Malik, and Christoph Feichtenhofer.
\newblock Multiscale vision transformers.
\newblock In {\em Proc. ICCV}, 2021.

\bibitem{feichtenhofer2020x3d}
Christoph Feichtenhofer.
\newblock {X3D}: Expanding architectures for efficient video recognition.
\newblock In {\em Proc. CVPR}, 2020.

\bibitem{Feichtenhofer2019}
Christoph Feichtenhofer, Haoqi Fan, Jitendra Malik, and Kaiming He.
\newblock {SlowFast} networks for video recognition.
\newblock In {\em Proc. ICCV}, 2019.

\bibitem{furnari2018leveraging}
Antonino Furnari, Sebastiano Battiato, and Giovanni Maria~Farinella.
\newblock Leveraging uncertainty to rethink loss functions and evaluation
  measures for egocentric action anticipation.
\newblock In {\em ECCV Workshops}, 2018.

\bibitem{furnari2020rolling}
Antonino Furnari and Giovanni Farinella.
\newblock Rolling-unrolling lstms for action anticipation from first-person
  video.
\newblock {\em PAMI}, 2020.

\bibitem{girdhar2019video}
Rohit Girdhar, Joao Carreira, Carl Doersch, and Andrew Zisserman.
\newblock Video action transformer network.
\newblock In {\em Proc. CVPR}, 2019.

\bibitem{girdhar2021anticipative}
Rohit Girdhar and Kristen Grauman.
\newblock {Anticipative Video Transformer}.
\newblock In {\em Proc. ICCV}, 2021.

\bibitem{girdhar2017actionvlad}
Rohit Girdhar, Deva Ramanan, Abhinav Gupta, Josef Sivic, and Bryan Russell.
\newblock {ActionVLAD}: Learning spatio-temporal aggregation for action
  classification.
\newblock In {\em Proc. CVPR}, 2017.

\bibitem{Goyal2017}
Priya Goyal, Piotr Doll{\'a}r, Ross Girshick, Pieter Noordhuis, Lukasz
  Wesolowski, Aapo Kyrola, Andrew Tulloch, Yangqing Jia, and Kaiming He.
\newblock {Accurate, large minibatch SGD: training ImageNet in 1 hour}.
\newblock {\em arXiv:1706.02677}, 2017.

\bibitem{Graham_2021_ICCV}
Benjamin Graham, Alaaeldin El-Nouby, Hugo Touvron, Pierre Stock, Armand Joulin,
  Herve Jegou, and Matthijs Douze.
\newblock {LeViT}: A vision transformer in {ConvNet}'s clothing for faster
  inference.
\newblock In {\em Proc. ICCV}, 2021.

\bibitem{Gu2018}
Chunhui Gu, Chen Sun, David~A. Ross, Carl Vondrick, Caroline Pantofaru, Yeqing
  Li, Sudheendra Vijayanarasimhan, George Toderici, Susanna Ricco, Rahul
  Sukthankar, Cordelia Schmid, and Jitendra Malik.
\newblock {AVA}: A video dataset of spatio-temporally localized atomic visual
  actions.
\newblock In {\em Proc. CVPR}, 2018.

\bibitem{hussein2019timeception}
Noureldien Hussein, Efstratios Gavves, and Arnold~WM Smeulders.
\newblock Timeception for complex action recognition.
\newblock In {\em Proc. CVPR}, 2019.

\bibitem{jiang2019stm}
Boyuan Jiang, MengMeng Wang, Weihao Gan, Wei Wu, and Junjie Yan.
\newblock {STM}: Spatiotemporal and motion encoding for action recognition.
\newblock In {\em Proc. CVPR}, 2019.

\bibitem{Kay2017}
Will Kay, Joao Carreira, Karen Simonyan, Brian Zhang, Chloe Hillier, Sudheendra
  Vijayanarasimhan, Fabio Viola, Tim Green, Trevor Back, Paul Natsev, et~al.
\newblock The kinetics human action video dataset.
\newblock {\em arXiv:1705.06950}, 2017.

\bibitem{klaser2008spatio}
Alexander Klaser, Marcin Marsza{\l}ek, and Cordelia Schmid.
\newblock A spatio-temporal descriptor based on 3d-gradients.
\newblock In {\em Proc. BMVC.}, 2008.

\bibitem{kondratyuk2021movinets}
Dan Kondratyuk, Liangzhe Yuan, Yandong Li, Li Zhang, Mingxing Tan, Matthew
  Brown, and Boqing Gong.
\newblock {MoViNets}: Mobile video networks for efficient video recognition.
\newblock In {\em Proc. CVPR}, 2021.

\bibitem{korbar2019scsampler}
Bruno Korbar, Du Tran, and Lorenzo Torresani.
\newblock Scsampler: Sampling salient clips from video for efficient action
  recognition.
\newblock In {\em Proc. ICCV}, 2019.

\bibitem{Laptev2008}
Ivan Laptev, Marcin Marszalek, Cordelia Schmid, and Benjamin Rozenfeld.
\newblock Learning realistic human actions from movies.
\newblock In {\em Proc. CVPR}, 2008.

\bibitem{lee2021video}
Sangmin Lee, Hak~Gu Kim, Dae~Hwi Choi, Hyung-Il Kim, and Yong~Man Ro.
\newblock Video prediction recalling long-term motion context via memory
  alignment learning.
\newblock In {\em Proc. CVPR}, 2021.

\bibitem{lee2018memory}
Sangho Lee, Jinyoung Sung, Youngjae Yu, and Gunhee Kim.
\newblock A memory network approach for story-based temporal summarization of
  360 videos.
\newblock In {\em Proc. CVPR}, 2018.

\bibitem{Li2018a}
Dong Li, Zhaofan Qiu, Qi Dai, Ting Yao, and Tao Mei.
\newblock Recurrent tubelet proposal and recognition networks for action
  detection.
\newblock In {\em Proc. ECCV}, 2018.

\bibitem{ego-exo}
Yanghao Li, Tushar Nagarajan, Bo Xiong, and Kristen Grauman.
\newblock Ego-exo: Transferring visual representations from third-person to
  first-person videos.
\newblock In {\em Proc. CVPR}, 2021.

\bibitem{li2021improved}
Yanghao Li, Chao-Yuan Wu, Haoqi Fan, Karttikeya Mangalam, Bo Xiong, Jitendra
  Malik, and Christoph Feichtenhofer.
\newblock Improved multiscale vision transformers for classification and
  detection.
\newblock {\em arXiv preprint arXiv:2112.01526}, 2021.

\bibitem{Li2018}
Zhenyang Li, Kirill Gavrilyuk, Efstratios Gavves, Mihir Jain, and Cees~GM
  Snoek.
\newblock {VideoLSTM} convolves, attends and flows for action recognition.
\newblock {\em Computer Vision and Image Understanding}, 2018.

\bibitem{lin2019tsm}
Ji Lin, Chuang Gan, and Song Han.
\newblock {TSM}: Temporal shift module for efficient video understanding.
\newblock In {\em Proc. ICCV}, 2019.

\bibitem{Lin2017}
Tsung-Yi Lin, Piotr Doll{\'a}r, Ross Girshick, Kaiming He, Bharath Hariharan,
  and Serge Belongie.
\newblock Feature pyramid networks for object detection.
\newblock In {\em Proc. CVPR}, 2017.

\bibitem{liu2018mobile}
Mason Liu and Menglong Zhu.
\newblock Mobile video object detection with temporally-aware feature maps.
\newblock In {\em Proc. CVPR}, 2018.

\bibitem{Swin}
Ze Liu, Yutong Lin, Yue Cao, Han Hu, Yixuan Wei, Zheng Zhang, Stephen Lin, and
  Baining Guo.
\newblock Swin transformer: Hierarchical vision transformer using shifted
  windows.
\newblock In {\em Proc. CVPR}, 2022.

\bibitem{loshchilov2017decoupled}
Ilya Loshchilov and Frank Hutter.
\newblock Decoupled weight decay regularization.
\newblock {\em arXiv preprint arXiv:1711.05101}, 2017.

\bibitem{neimark2021video}
Daniel Neimark, Omri Bar, Maya Zohar, and Dotan Asselmann.
\newblock Video transformer network.
\newblock {\em arXiv preprint arXiv:2102.00719}, 2021.

\bibitem{Ng2015}
Joe Yue-Hei Ng, Matthew Hausknecht, Sudheendra Vijayanarasimhan, Oriol Vinyals,
  Rajat Monga, and George Toderici.
\newblock Beyond short snippets: Deep networks for video classification.
\newblock In {\em Proc. CVPR}, 2015.

\bibitem{pan2021actor}
Junting Pan, Siyu Chen, Mike~Zheng Shou, Yu Liu, Jing Shao, and Hongsheng Li.
\newblock Actor-context-actor relation network for spatio-temporal action
  localization.
\newblock In {\em Proc. CVPR}, 2021.

\bibitem{MotionFormer}
Mandela Patrick, Dylan Campbell, Yuki~M Asano, Ishan Misra~Florian Metze,
  Christoph Feichtenhofer, Andrea Vedaldi, Jo Henriques, et~al.
\newblock Keeping your eye on the ball: Trajectory attention in video
  transformers.
\newblock In {\em NeurIPS}, 2021.

\bibitem{Peng2014}
Xiaojiang Peng, Changqing Zou, Yu Qiao, and Qiang Peng.
\newblock Action recognition with stacked fisher vectors.
\newblock In {\em Proc. ECCV}, 2014.

\bibitem{Qiu2017}
Zhaofan Qiu, Ting Yao, and Tao Mei.
\newblock Learning spatio-temporal representation with pseudo-3d residual
  networks.
\newblock In {\em Proc. ICCV}, 2017.

\bibitem{ilija_2020}
Ilija Radosavovic, Raj Prateek~Kosaraju, Ross Girshick, Kaiming He, and Piotr
  Dollár.
\newblock Designing network design spaces.
\newblock In {\em Proc. CVPR}, 2020.

\bibitem{rae2019compressive}
Jack~W Rae, Anna Potapenko, Siddhant~M Jayakumar, and Timothy~P Lillicrap.
\newblock Compressive transformers for long-range sequence modelling.
\newblock In {\em ICLR}, 2019.

\bibitem{rae2020transformers}
Jack~W Rae and Ali Razavi.
\newblock Do transformers need deep long-range memory.
\newblock In {\em ACL}, 2020.

\bibitem{Ren2015}
Shaoqing Ren, Kaiming He, Ross Girshick, and Jian Sun.
\newblock {Faster R-CNN}: Towards real-time object detection with region
  proposal networks.
\newblock In {\em NIPS}, 2015.

\bibitem{sener2021technical}
Fadime Sener, Dibyadip Chatterjee, and Angela Yao.
\newblock Technical report: Temporal aggregate representations.
\newblock {\em arXiv preprint arXiv:2106.03152}, 2021.

\bibitem{shaw-etal-2018-self}
Peter Shaw, Jakob Uszkoreit, and Ashish Vaswani.
\newblock Self-attention with relative position representations.
\newblock In {\em NAACL}, 2018.

\bibitem{sukhbaatar2019adaptive}
Sainbayar Sukhbaatar, Edouard Grave, Piotr Bojanowski, and Armand Joulin.
\newblock Adaptive attention span in transformers.
\newblock In {\em ACL}, 2019.

\bibitem{sukhbaatar2021not}
Sainbayar Sukhbaatar, Da Ju, Spencer Poff, Stephen Roller, Arthur Szlam, Jason
  Weston, and Angela Fan.
\newblock Not all memories are created equal: Learning to forget by expiring.
\newblock In {\em ICML}, 2021.

\bibitem{Sun2017}
Lin Sun, Kui Jia, Kevin Chen, Dit-Yan Yeung, Bertram~E Shi, and Silvio
  Savarese.
\newblock Lattice long short-term memory for human action recognition.
\newblock In {\em Proc. ICCV}, 2017.

\bibitem{tan2020equalization}
Jingru Tan, Changbao Wang, Buyu Li, Quanquan Li, Wanli Ouyang, Changqing Yin,
  and Junjie Yan.
\newblock Equalization loss for long-tailed object recognition.
\newblock In {\em Proc. CVPR}, 2020.

\bibitem{tan20201st}
Jingru Tan, Gang Zhang, Hanming Deng, Changbao Wang, Lewei Lu, Quanquan Li, and
  Jifeng Dai.
\newblock 1st place solution of lvis challenge 2020: A good box is not a
  guarantee of a good mask.
\newblock {\em arXiv preprint arXiv:2009.01559}, 2020.

\bibitem{tang2020asynchronous}
Jiajun Tang, Jin Xia, Xinzhi Mu, Bo Pang, and Cewu Lu.
\newblock Asynchronous interaction aggregation for action detection.
\newblock In {\em Proc. ECCV}, 2020.

\bibitem{Taylor2010}
Graham~W Taylor, Rob Fergus, Yann LeCun, and Christoph Bregler.
\newblock Convolutional learning of spatio-temporal features.
\newblock In {\em Proc. ECCV}, 2010.

\bibitem{deit}
Hugo Touvron, Matthieu Cord, Matthijs Douze, Francisco Massa, Alexandre
  Sablayrolles, and Herve Jegou.
\newblock Training data-efficient image transformers and distillation through
  attention.
\newblock In {\em Proc. ICML}, 2021.

\bibitem{touvron2021going}
Hugo Touvron, Matthieu Cord, Alexandre Sablayrolles, Gabriel Synnaeve, and
  Herv{\'e} J{\'e}gou.
\newblock Going deeper with image transformers.
\newblock In {\em Proc. ICCV}, 2021.

\bibitem{Tran2015}
Du Tran, Lubomir Bourdev, Rob Fergus, Lorenzo Torresani, and Manohar Paluri.
\newblock Learning spatiotemporal features with {3D} convolutional networks.
\newblock In {\em Proc. ICCV}, 2015.

\bibitem{Tran2019}
Du Tran, Heng Wang, Lorenzo Torresani, and Matt Feiszli.
\newblock Video classification with channel-separated convolutional networks.
\newblock In {\em Proc. ICCV}, 2019.

\bibitem{vaswani2017attention}
Ashish Vaswani, Noam Shazeer, Niki Parmar, Jakob Uszkoreit, Llion Jones,
  Aidan~N Gomez, Lukasz Kaiser, and Illia Polosukhin.
\newblock Attention is all you need.
\newblock In {\em NeurIPS}, 2017.

\bibitem{Wang2013a}
Heng Wang, Alexander Kl{\"a}ser, Cordelia Schmid, and Cheng-Lin Liu.
\newblock Dense trajectories and motion boundary descriptors for action
  recognition.
\newblock {\em IJCV}, 2013.

\bibitem{Wang2013}
Heng Wang and Cordelia Schmid.
\newblock Action recognition with improved trajectories.
\newblock In {\em Proc. ICCV}, 2013.

\bibitem{wang2009evaluation}
Heng Wang, Muhammad~Muneeb Ullah, Alexander Klaser, Ivan Laptev, and Cordelia
  Schmid.
\newblock Evaluation of local spatio-temporal features for action recognition.
\newblock In {\em BMVC}, 2009.

\bibitem{Wang2015}
Limin Wang, Yu Qiao, and Xiaoou Tang.
\newblock Action recognition with trajectory-pooled deep-convolutional
  descriptors.
\newblock In {\em Proc. CVPR}, 2015.

\bibitem{Wang2016a}
Limin Wang, Yuanjun Xiong, Zhe Wang, Yu Qiao, Dahua Lin, Xiaoou Tang, and Luc
  {Val Gool}.
\newblock Temporal segment networks: Towards good practices for deep action
  recognition.
\newblock In {\em Proc. ECCV}, 2016.

\bibitem{wang2021pyramid}
Wenhai Wang, Enze Xie, Xiang Li, Deng-Ping Fan, Kaitao Song, Ding Liang, Tong
  Lu, Ping Luo, and Ling Shao.
\newblock Pyramid vision transformer: A versatile backbone for dense prediction
  without convolutions.
\newblock In {\em Proc. ICCV}, 2021.

\bibitem{Wang2018}
Xiaolong Wang, Ross Girshick, Abhinav Gupta, and Kaiming He.
\newblock Non-local neural networks.
\newblock In {\em Proc. CVPR}, 2018.

\bibitem{wang2021interactive}
Xiaohan Wang, Linchao Zhu, Heng Wang, and Yi Yang.
\newblock Interactive prototype learning for egocentric action recognition.
\newblock In {\em Proc. ICCV}, 2021.

\bibitem{Wu2019}
Chao-Yuan Wu, Christoph Feichtenhofer, Haoqi Fan, Kaiming He, Philipp
  Kr\"{a}henb\"{u}hl, and Ross Girshick.
\newblock Long-term feature banks for detailed video understanding.
\newblock In {\em Proc. CVPR}, 2019.

\bibitem{wu2021towards}
Chao-Yuan Wu and Philipp Krahenbuhl.
\newblock Towards long-form video understanding.
\newblock In {\em Proc. CVPR}, 2021.

\bibitem{Wu2018}
Chao-Yuan Wu, Manzil Zaheer, Hexiang Hu, R Manmatha, Alexander~J Smola, and
  Philipp Kr{\"a}henb{\"u}hl.
\newblock Compressed video action recognition.
\newblock In {\em Proc. CVPR}, 2018.

\bibitem{Xie2017}
Saining Xie, Ross Girshick, Piotr Doll{\'a}r, Zhuowen Tu, and Kaiming He.
\newblock Aggregated residual transformations for deep neural networks.
\newblock In {\em Proc. CVPR}, 2017.

\bibitem{Xie2018}
Saining Xie, Chen Sun, Jonathan Huang, Zhuowen Tu, and Kevin Murphy.
\newblock Rethinking spatiotemporal feature learning for video understanding.
\newblock {\em arXiv:1712.04851}, 2017.

\bibitem{yuan2021tokens}
Li Yuan, Yunpeng Chen, Tao Wang, Weihao Yu, Yujun Shi, Francis~EH Tay, Jiashi
  Feng, and Shuicheng Yan.
\newblock Tokens-to-token {ViT}: Training vision transformers from scratch on
  imagenet.
\newblock In {\em Proc. ICCV}, 2021.

\bibitem{yue2015beyond}
Joe Yue-Hei~Ng, Matthew Hausknecht, Sudheendra Vijayanarasimhan, Oriol Vinyals,
  Rajat Monga, and George Toderici.
\newblock Beyond short snippets: Deep networks for video classification.
\newblock In {\em Proc. CVPR}, 2015.

\bibitem{Zhou2017}
Bolei Zhou, Alex Andonian, Aude Oliva, and Antonio Torralba.
\newblock Temporal relational reasoning in videos.
\newblock In {\em ECCV}, 2018.

\bibitem{zhu17fgfa}
Xizhou Zhu, Yujie Wang, Jifeng Dai, Lu Yuan, and Yichen Wei.
\newblock Flow-guided feature aggregation for video object detection.
\newblock In {\em Proc. ICCV}, 2017.

\bibitem{Zolfaghari2018}
Mohammadreza Zolfaghari, Kamaljeet Singh, and Thomas Brox.
\newblock {ECO:} efficient convolutional network for online video
  understanding.
\newblock In {\em Proc. ECCV}, 2018.

\end{thebibliography}
	}
	
\end{document}